\documentclass[10pt,twocolumn,letterpaper]{article}

\usepackage[pagenumbers]{cvpr} 

\usepackage{graphicx}
\usepackage{amsmath}
\usepackage{amssymb}
\usepackage{booktabs}
\usepackage{multirow}
\usepackage[accsupp]{axessibility}

\usepackage[pagebackref,breaklinks,colorlinks]{hyperref}

\newcommand\model{UNITS\xspace}

\usepackage[capitalize]{cleveref}
\crefname{section}{Sec.}{Secs.}
\Crefname{section}{Section}{Sections}
\Crefname{table}{Table}{Tables}
\crefname{table}{Tab.}{Tabs.}

\def\cvprPaperID{10651} 
\def\confName{CVPR}
\def\confYear{2023}

\begin{document}

\title{Towards Unified Scene Text Spotting based on Sequence Generation}

\author{
Taeho Kil\textsuperscript{\rm 1}\thanks{Corresponding author.}\hspace{0.4cm}
Seonghyeon Kim\textsuperscript{\rm 2}\thanks{This work was done while the authors were at Naver Cloud.}\hspace{0.4cm}
Sukmin Seo\textsuperscript{\rm 1}\hspace{0.4cm}
Yoonsik Kim\textsuperscript{\rm 1}\hspace{0.4cm}
Daehee Kim\textsuperscript{\rm 1}\vspace{0.25cm}\\
\textsuperscript{\rm 1}Naver Cloud\hspace{0.4cm}
\textsuperscript{\rm 2}Kakao Brain\vspace{0.1cm}\\
{\tt\small \{taeho.kil, sukmin.seo, yoonsik.kim90, daehee.k\}@navercorp.com, matt.mldev@kakaobrain.com}}

\maketitle

\begin{abstract}
Sequence generation models have recently made significant progress in unifying various vision tasks.
Although some auto-regressive models have demonstrated promising results in end-to-end text spotting, they use specific detection formats while ignoring various text shapes and are limited in the maximum number of text instances that can be detected.
To overcome these limitations, we propose a UNIfied scene Text Spotter, called~\model.
Our model unifies various detection formats, including quadrilaterals and polygons, allowing it to detect text in arbitrary shapes. 
Additionally, we apply starting-point prompting to enable the model to extract texts from an arbitrary starting point, thereby extracting more texts beyond the number of instances it was trained on.
Experimental results demonstrate that our method achieves competitive performance compared to state-of-the-art methods.
Further analysis shows that~\model can extract a larger number of texts than it was trained on.
We provide the code for our method at \href{https://github.com/clovaai/units}{https://github.com/clovaai/units}.
\end{abstract}

\section{Introduction}
End-to-end scene text spotting, which can jointly detect and recognize text from an image at once, has recently gained significant attention.
This work has practical applications in various fields such as visual navigation, visual question answering, and document image understanding.

In this paper, we formulate scene text spotting as a sequence generation task.
This approach casts the vision task as a language modeling task conditioned on the image and text prompt~\cite{chen2021pix2seq, yang2022unitab, lu2022unified, chen2022unified}.
By formulating the output as a sequence of discrete tokens, the vision task can be performed by generating a sequence through an auto-regressive transformer decoder.
Recent studies have attempted to integrate various tasks into an auto-regressive model, including text spotting.
SPTS~\cite{peng2022spts} treated all detection formats as the single central point and predicted the coordinate tokens of the central point and word tokens auto-regressively.
However, there are several challenges associated with applying the sequence generation approach directly to scene text spotting.

\begin{figure}
  \centering
  \begin{subfigure}{.49\linewidth}
    \includegraphics[width=1.0\linewidth]{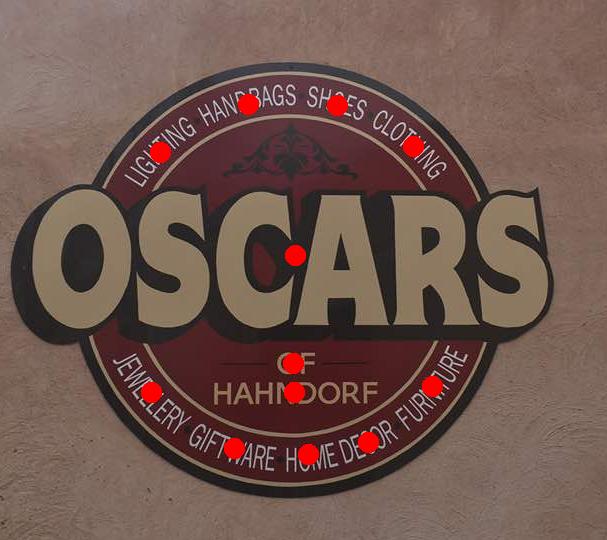}
    \caption{Single central point format.}
  \end{subfigure}
  \begin{subfigure}{.49\linewidth}
    \includegraphics[width=1.0\linewidth]{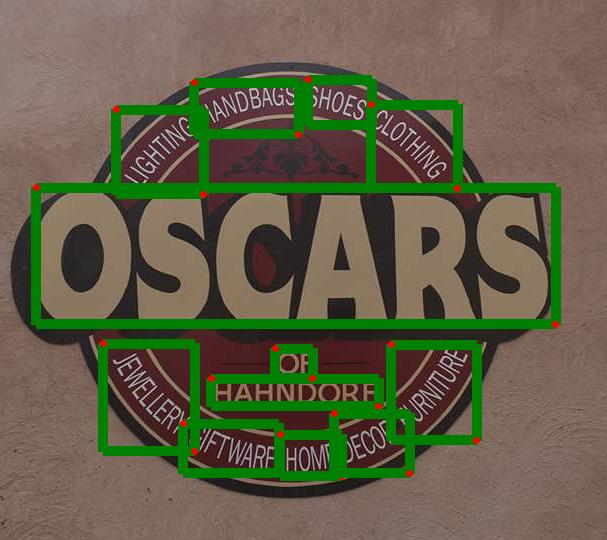}
    \caption{Bounding box format.}
  \end{subfigure}
  \begin{subfigure}{.49\linewidth}
    \includegraphics[width=1.0\linewidth]{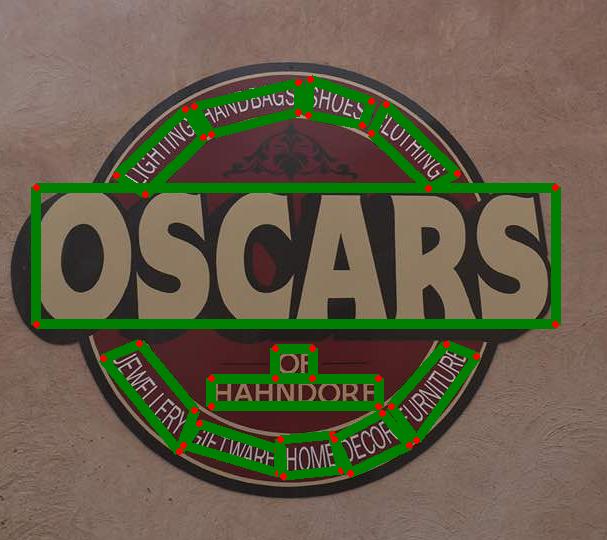}
    \caption{Quadrilateral format.}
  \end{subfigure}
  \begin{subfigure}{.49\linewidth}
    \includegraphics[width=1.0\linewidth]{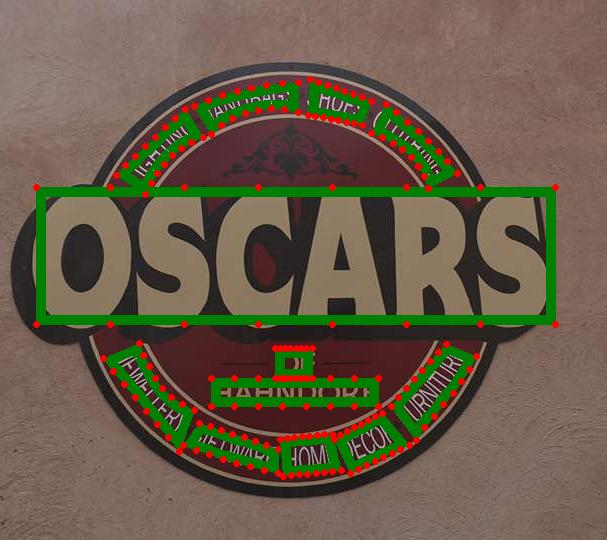}
    \caption{Polygonal format.}
  \end{subfigure}
  \caption{Various types of detection formats. The green line represents the boundary shape of the detection format, and the red dot represents the points used for the corresponding format.}
\label{fig:detection_formats}
\end{figure}

\begin{figure}
  \centering
  \begin{subfigure}{.49\linewidth}
    \includegraphics[width=1.0\linewidth]{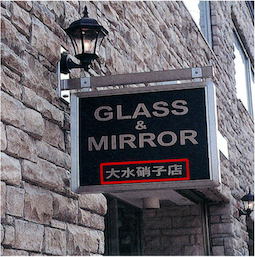}
    \caption{Original image.}
  \end{subfigure}
  \begin{subfigure}{.49\linewidth}
    \includegraphics[width=1.0\linewidth]{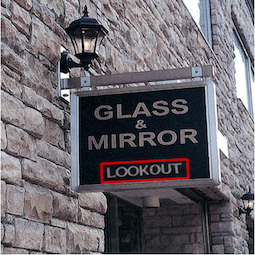}
    \caption{Result of image translation.}
  \end{subfigure}
  \caption{Example of image translation using scene text editing.}
\label{fig:applications}
\end{figure}

As illustrated in~\cref{fig:detection_formats}, there are various methods to indicate the location and boundaries of text instances, and each method has its own trade-offs in terms of annotation costs and potential applicability.
For instance, a single central point or bounding box annotation has a relatively low annotation cost and is appropriate when the detailed shape information is not necessary.
However, in many fields where text spotting is applied, handling text location information as only a single point might be insufficient.
As shown in~\cref{fig:applications}, in the scene text editing~\cite{roy2020stefann, lee2021rewritenet}, which converts the text in the image to desired text while preserving the text style, detailed text shape extraction is required.
Therefore, for such tasks, outputting text location information in quadrilateral or polygonal formats beyond a single point is necessary.
Similarly, in visual document understanding, which utilizes optical character recognition as a pre-processing step, more detailed detection information of texts could be useful.
In summary, since each detection format has its own trade-offs, it is better to cover all detection formats instead of relying on only one.

Also, conventional sequence generation models have a limitation in terms of generating sequences longer than the maximum length the model has been trained on.
This limitation is particularly problematic for scene text spotting, as it requires generating more points than bounding boxes and generating text transcriptions in addition to object classes.
For text-rich images such as documents, this limitation becomes a bottleneck in extracting all the texts.

To address these challenges in applying the sequence generation method to text spotting, we propose a novel end-to-end UNIfied scene Text Spotter, called~\model, which aims to overcome these limitations.
We unify various detection formats, such as quadrilateral and polygon, into the model, allowing it to detect arbitrary-shaped text areas.
This approach enables the model to learn different annotation formats together using a single unified model, and it can extract all detection formats.
We use prompts to unify various detection formats, enabling a single model to predict the coarse or fine-grained location information of text instances.

Moreover, we introduce starting-point prompting, which enables the model to extract texts from arbitrary starting points, allowing it to generate longer sequences than the maximum decoding length.
This approach enhances the model's efficiency and enables it to extract more text than the number of text boxes it has been trained on.

In addition, we employ a multi-way transformer decoder from the mixture-of-experts (MoE)~\cite{fedus2021switch,wang2021vlmo,bao2022vl}.
This decoder separates each time stamp into detection and recognition, allowing the model to converge faster by guiding the model to generate a detection or a recognition token and assigning an expert accordingly.
Each block of the multi-way transformer consists of shared attention modules and two task experts: detection and recognition experts.

Experimental results demonstrate that our proposed method achieves competitive performance on text spotting benchmarks while extracting texts in various detection formats using a single unified model.

Our main contributions are:

\begin{itemize}
  \item We propose a novel sequence generation-based scene text spotting method that can extract arbitrary-shaped text areas by unifying various detection formats.
  \vspace{-0.5em}  
  \item Our model can extract more texts than the decoder length allows using the starting-point prompt, which generalizes the model to spot more texts than it has been trained on.
  \vspace{-0.5em}  
  \item Experimental results demonstrate that our method achieves competitive performance on text spotting benchmarks and provides additional functionalities
\end{itemize}

\section{Related Work}
\begin{figure*}
  \centering
  \includegraphics[width=1.0\linewidth]{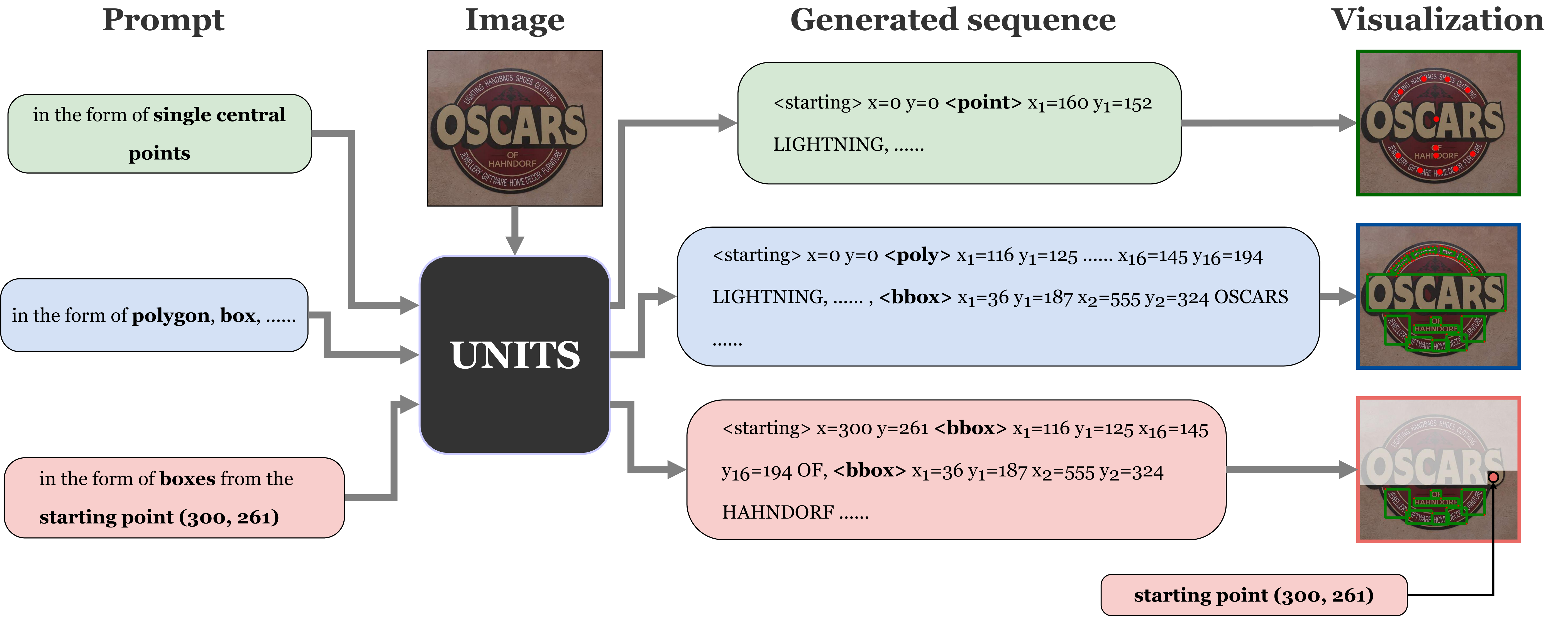}
  \caption{Pipeline of the proposed text spotting method. The information of each instance in the sequence consists of the detection format token, coordinate tokens for detection, and text transcriptions. Coordinate tokens are generated according to the desired detection format. The model sequentially detects and recognizes only text instances located after a specific location determined by the starting-point prompt.}
  \label{fig:pipeline}
\end{figure*}

\noindent\textbf{Text Spotting}.
The text spotting that detects and recognizes text at once has recently attracted a lot of attention.
Many studies improved the performance by learning the detector and recognizer simultaneously~\cite{liu2020abcnet, liao2020mask, peng2022spts, wang2021pgnet, liu2018fots, kittenplon2022towards, kim2022deer, baek2020character, feng2019textdragon, qiao2021mango, zhang2022text, huang2022swintextspotter}, rather than separately~\cite{liao2018textboxes++, liao2017textboxes, shi2016end, borisyuk2018rosetta}.
Rosetta++~\cite{borisyuk2018rosetta} detected text in the form of a bounding box.
Some methods~\cite{liao2018textboxes++, zhou2017east, liu2018fots} treated text location information with the quadrilateral format and detected inclined text beyond horizontal text.
For arbitrary-shaped text detection, some works used parametric Bezier curve representations~\cite{liu2020abcnet, liu2021abcnet, zhang2022text}.
The segmentation-based method~\cite{liao2020mask, qiao2021mango, liao2020real, qiao2021mango} can handle all text instances of various shapes, however, there is a disadvantage that it needs complex post-processing such as connected components filtering. 
Some methods~\cite{baek2019character, baek2020character, wang2021pgnet, liao2020mask, feng2019textdragon, qiao2021mango} used character-level as well as word-level annotations. It enables the detection of arbitrarily-shaped texts, however, character-level annotation is expensive and it requires post-processing like word grouping.

Recently, there were also attempts to develop a model that can deal with various detection formats.
DEER~\cite{kim2022deer} recognized words by guiding text instance location as a simple single reference point without RoI cropping.
TTS~\cite{kittenplon2022towards} proposed a weakly supervised learning method that makes the box, and mask predictors lightweight and treated various types of annotation formats while sharing the backbone and encoder. 
SwinTextSpotter~\cite{huang2022swintextspotter} and MaskTextSpotter~\cite{liao2020mask} also used separated task-specific heads corresponding to different annotation formats.
Although the backbone and encoder are shared in these works, the task-specific heads are separated by detection formats so the proposed method is a more strictly unified approach.
The proposed method can learn different annotation types at once and extract the location information of the texts into the desired detection formats.

\noindent\textbf{Sequence Generation-based Approach}.
Recently, for vision tasks, the methods to apply the seq2seq approach used in NLP have been studied. 
Instead of designing a model specific to the vision task, this approach uses a simple transformer encoder and decoder structure and performs the task by predicting sequence.
Pix2seq~\cite{chen2021pix2seq} was the first attempt at this approach, it showed that objects can be detected well even in random order without any matching algorithm like Hungarian matching.
UniTAB~\cite{yang2022unitab} performed multi-task learning of vision language tasks such as object detection, captioning, and visual question answering by utilizing the method of guiding tasks through task prompts like T5~\cite{raffel2020exploring}.
Unified-IO~\cite{lu2022unified} further jointly trained various vision language tasks including image generation by using the unified input/output formats for all tasks.
The generation-based methods can unify various tasks and modalities in simple transformer encoder-decoder architecture.

There was a study in which a sequence generation-based method has been applied for text spotting. 
SPTS~\cite{peng2022spts} treated all location information as the central point and predicted the coordinate tokens of the central point and word transcription tokens auto-regressively.
This method showed that single point location and transcription for text instances are well generated, however, it can not handle the more complex detection formats.
Also, it has the limitations about decoder length that common sequence generation methods have. 
This method cannot handle the case where the number of objects is larger than the number of objects allowed by the decoder length.
The proposed method can handle the various types of detection formats through prompts and can extract texts beyond the number of text instances that have been trained.

\section{Method}
We propose an end-to-end scene text spotting method, named~\model, which is based on a unified interface that handles various text detection formats and determines the starting point of text to extract.
The overall pipeline is shown in~\cref{fig:pipeline}.
The details of the unified interface are described in~\ref{subsec:interface}, while the architecture of the proposed method, including the multi-way transformer decoder and objective function, is presented in~\ref{subsec:arch}.

\subsection{Unified Interface for Text Spotting}
\label{subsec:interface}

The sequence construction method of~\model follows recent methods~\cite{chen2021pix2seq, chen2022unified, lu2022unified, yang2022unitab, peng2022spts}.
The location and transcription of text instances are converted into discrete tokens through tokenization and concatenated in each instance.
As the location token, the \(x, y\) coordinates of the point are normalized to the width and height of the image, and then quantized to $[0, n_{bins}]$. The bin size $n_{bins}$ means the number of coordinate vocab, we set the bin size as 1,000.

Given the significant differences in detection formats, existing methods use task-specific heads.
To handle various format types with only a single model, we introduce a new sequence interface by adding prompts for each text instance.
Before generating the output of each text instance, the model determines the shape of the text area to be extracted through a prompt.
This prompt named detection format token is fed as input, and the model can extract the text instance in the desired format corresponding to the prompt.
As shown in~\cref{fig:pipeline}, when predicting texts in the form of the polygon, we insert the detection format token $<$poly$>$ before location tokens, and when treating bounding box-shaped instances, we insert the detection format token $<$bbox$>$.
After the detection format token, location information is extracted according to this format, and text transcription is generated next.

If there are more text instances in the image than the number of instances determined by the decoder length, it can cause a problem where the sequence generation model cannot detect and recognize all texts in the image.
To overcome this problem, we introduce starting-point prompting, which can guide the image region where text should be detected and can extract texts beyond the number of text instances that the model has been trained on.

\noindent\textbf{Detection Format Token}.
The use of detection format tokens enables a single model to handle multiple detection formats using a shared vocabulary.
The special tokens and the resulting output sequence are illustrated in~\cref{fig:interface}.
There are five texts, and they are arranged in a 1D sequence using raster scan order as `OLD', `MILL', `RECEPTIO', `HOTEL', and `MOTEL'.
For each text instance, the detection format token determines whether it should be represented as a bounding box or another format, such as a polygon or a single point.
In the second turn, the detection format token corresponding to the box annotation is used, resulting in the extraction of the `MILL' instance as a bounding box.

In this paper, we use a total of four detection format tokens: central point, bounding box, quadrilateral, and polygonal formats.
The central point, bounding box, quadrilateral, and polygonal formats are expressed as 1, 2, 4, and 16 points, respectively, and the length of transcriptions is fixed at 25.
When constructing transcription tokens, if the text length exceeds 25, the excess part is removed, while if the text length is shorter, it is padded.
Through this sequence construction configuration, we propose a unified interface that can handle multiple detection formats simultaneously.

\noindent\textbf{Starting-Point Prompting}.
Multiple text instances are generated in raster scan order, and sequence generation is guided by the starting-point prompt.
Thus,~\model can read texts from a specific location in raster scan order by giving a specific starting point.
This makes it possible to detect a large number of texts, such as texts in a document, even with a limited decoder length.

\begin{figure}
  \centering
  \begin{subfigure}{1.0\linewidth}
    \includegraphics[width=1.0\linewidth]{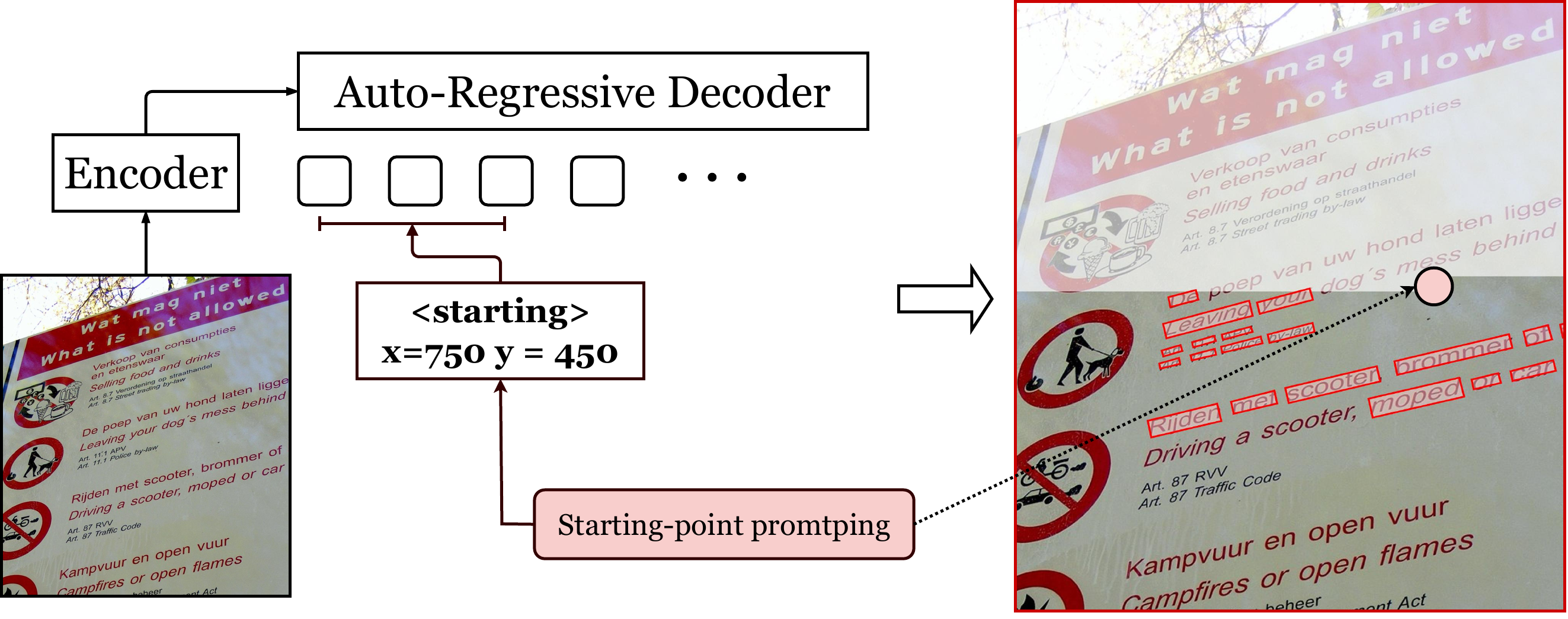}
    \caption{Training step.}
  \end{subfigure}
  \hfill
  \begin{subfigure}{1.0\linewidth}
    \includegraphics[width=1.0\linewidth]{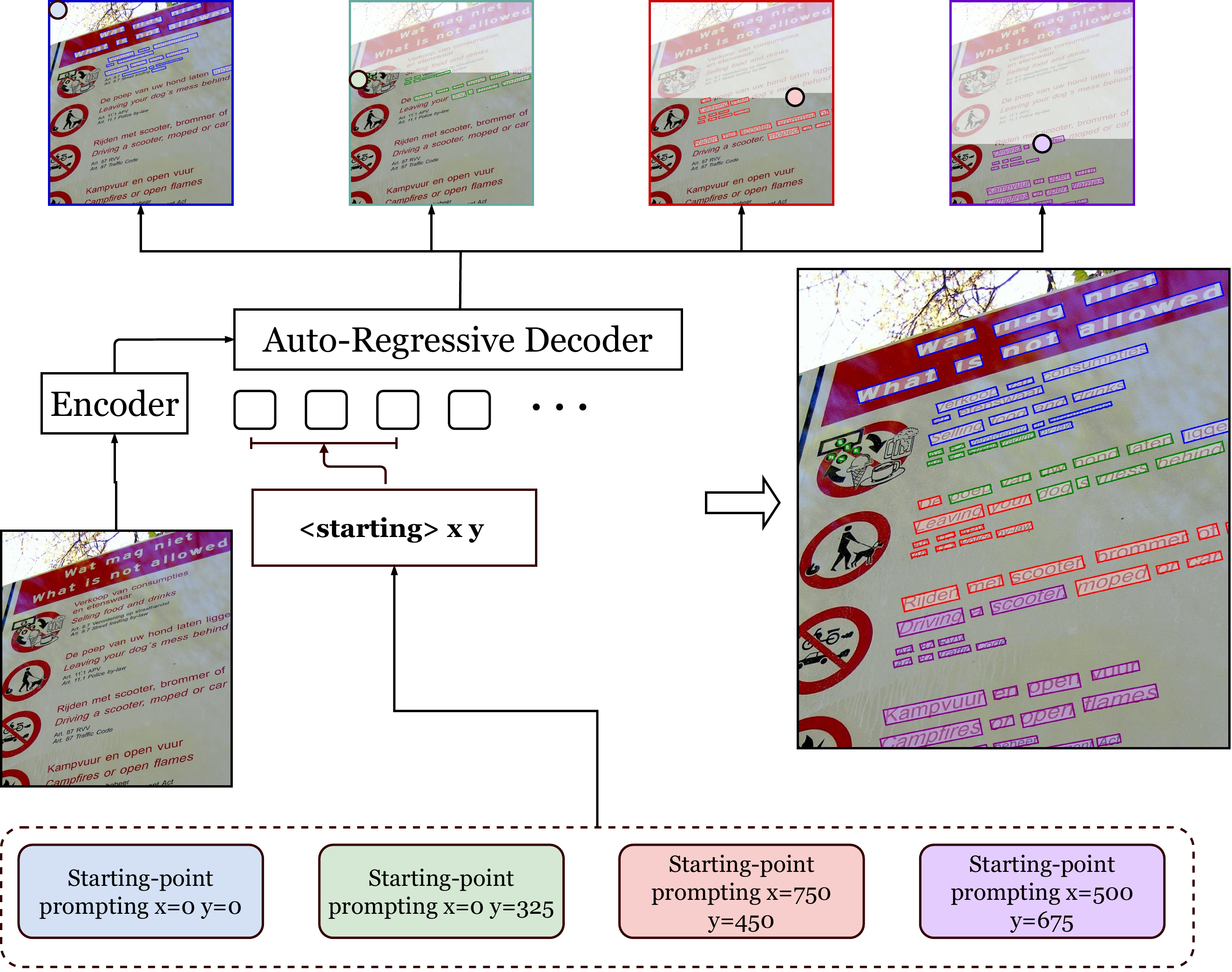}
    \hfill
    \caption{Testing step.}
  \end{subfigure}
  \caption{Illustration of starting-point prompting. In the training step, the model learns the ability to extract texts after a randomly selected starting point. In the testing step, the model extracts all texts by using multiple prompts.}
  \label{fig:location_prompting}
\end{figure}

As shown in~\cref{fig:location_prompting}, only texts located after the starting point are detected and recognized in raster scan order.
This strategy teaches the model to detect text instances from an arbitrary starting point while ignoring the text instances before the starting point.
The starting-point prompt composes of the \(x, y\) coordinates of the point.
When testing, first we set the starting point as the top-left point of the image, then a sequence of text instances in front is first generated.
If the generated output sequence does not end with an $<$eos$>$ token, our method sets the starting point as the last detected text position in the previous step, then continues and enables the re-generation of a text sequence corresponding to the remaining objects.
By repeating this process until $<$eos$>$ token appears, the model is able to cover whole images in raster scan order, and all texts in the image can be detected and recognized by concatenating parts of texts.
In the training step, the starting point in the image is set randomly so that the model learns the ability to detect text instances from the arbitrary starting point in the image.

\subsection{Architecture and Objective}
\label{subsec:arch}

\noindent\textbf{Architecture}.
Our model follows a basic encoder-decoder structure, where the text spotting task is cast as a language modeling task conditioned on the image, similar to conventional approaches~\cite{chen2021pix2seq, chen2022unified, lu2022unified, yang2022unitab, peng2022spts}.
For encoding the raw image, we use the Swin Transformer~\cite{liu2021swin} as the image encoder.
To generate the text sequence, we use a Transformer decoder that generates one token at a time conditioned on the previous tokens and encoded image features.

\begin{figure}
  \centering
  \includegraphics[width=1.0\linewidth]{figures/interface.pdf}
  \caption{Illustration of unified interface for text spotting.}
  \label{fig:interface}
\end{figure}

\noindent\textbf{Multi-way Decoder}.
We use a multi-way transformer decoder to handle the learning of multiple detection formats simultaneously.
Since each detection format has a different number of location tokens, learning multiple formats with varying patterns in the sequence is a more challenging task than learning a single format with a fixed pattern.
The multi-way decoder uses strong supervision by indicating whether it is time to generate detection or recognition tokens, thus helping to alleviate the difficulty of our model.

Each block of the multi-way transformer decoder consists of shared attention modules and two task experts: the detection expert and the recognition expert.
When generating the sequence, our model generates detection information for text instances first, followed by transcription information.
This allows the two tasks to be separated by the time stamp of the sequence. Each feedforward neural network (FFN) expert helps the single unified model to focus on detection and recognition, respectively.

\noindent\textbf{Objective}.
Since~\model predicts tokens through a decoder without a task-specific head, we train the model with a standard cross-entropy loss.
At each time stamp \(j\), the model aims to maximize the likelihood of the target tokens \(\tilde{y}\) given the input image \(I\) and previously generated tokens \(y_{1:j-1}\):
\begin{equation}
  L = - \sum_{j=1}^{N} w_{j} \log P(\tilde{y}_{j} \mid I, y_{1:j-1}),
\end{equation}
where \(y\) and \(\tilde{y}\) denote the input and target sequence, respectively, and \(N\) is the sequence length of sequence.
Also, \(w_j\) is the weight value for the \(j\)-th token, 0 is assigned at prompt token to exclude it from the loss calculation, and 1 is assigned for the remaining tokens.

\section{Experiments}
{
\begin{table*}
\centering
\begin{tabular}{lccccccc}
\toprule
\multirow{2}{*}{Method}                 & \multicolumn{3}{c}{Detection} & \multicolumn{4}{c}{End-to-End}    \\
\cmidrule(lr){2-4} \cmidrule(lr){5-8}
                                        & Recall & Precision & F-measure & Strong & Weak & Generic & None \\
\midrule
CRAFTS~\cite{baek2020character}         & 85.3  & 89.0  & 87.1  & 83.1  & 82.1  & 74.9  & -     \\
MaskTextSpotter v3~\cite{liao2020mask}  & -     & -     & -     & 83.3  & 78.1  & 74.2  & -     \\
ABCNet v2~\cite{liu2020abcnet}          & 86.0  & 90.4  & 88.1  & 82.7  & 78.5  & 73.0  & -     \\
MANGO~\cite{qiao2021mango}              & -     & -     & -     & 85.4  & 80.1  & 73.9  & -     \\
DEER~\cite{kim2022deer}                 & 86.2  & \underline{93.7}  & 89.8  & 82.7  & 79.1  & 75.6  & 71.7  \\
SwinTextSpotter~\cite{huang2022swintextspotter} & -     & -     & -     & 83.9  & 77.3  & 70.5  & -     \\
TESTR~\cite{zhang2022text}              & 89.7  & 90.3  & 90.0  & 85.2  & 79.4  & 73.6  & 65.3     \\
TTS~\cite{kittenplon2022towards}        & -     & -     & -     & 85.2  & 81.7  & 77.4  & -     \\
GLASS~\cite{ronen2022glass}             & -     & -     & -     & 84.7  & 80.1  & 76.3  & -     \\
\midrule
UNITS$_\textnormal{Shared}$             & \underline{90.5}  & 93.6  & \underline{92.0}  & \underline{88.4}  & \underline{83.9}  & \underline{79.7}  & \underline{78.5}  \\
UNITS                                   & \textbf{91.0}  & \textbf{94.0}  & \textbf{92.5}  & \textbf{89.0}  & \textbf{84.1}  & \textbf{80.3}  & \textbf{78.7}  \\
\bottomrule
\end{tabular}
\caption{Experiment results on ICDAR 2015. ``Strong'', ``Weak'', ``Generic'', and ``None'' represent recognition with each lexicon respectively.}
\label{tab:exp_ic15}
\end{table*}
}
{

\begin{table}
\centering
\begin{tabular}{lccccc}
\toprule
\multirow{2}{*}{Method}                 & \multicolumn{1}{c}{Detection} & \multicolumn{2}{c}{End-to-End}    \\
\cmidrule(lr){2-2} \cmidrule(lr){3-4}
                                        & F-measure & None & Full \\
\midrule
CRAFTS~\cite{baek2020character}         & 87.4  & \textbf{78.7}  & -     \\
MaskTextSpotter v3~\cite{liao2020mask}  & -     & 71.2  & 78.4  \\
ABCNet v2~\cite{liu2020abcnet}          & 87.0  & 70.4  & 78.1  \\
MANGO~\cite{qiao2021mango}              & -     & 72.9  & 83.6  \\
DEER~\cite{kim2022deer}                 & 85.7  & 74.8  & 83.3  \\
SwinTextSpotter~\cite{huang2022swintextspotter}& 88.0  & 74.3  & 84.1  \\
TESTR~\cite{zhang2022text}              & 86.9  & 73.3  & 83.9  \\
TTS~\cite{kittenplon2022towards}        & -     & 75.6  & 84.4  \\
GLASS~\cite{ronen2022glass}             & -     & 76.6  & 83.0  \\
\midrule
UNITS$_\textnormal{Shared}$             & \underline{88.4}  & 77.3  & \underline{85.0}  \\
UNITS                                   & \textbf{89.8}  & \textbf{78.7}  & \textbf{86.0}  \\
\bottomrule
\end{tabular}
\caption{Experiment results on Total-Text. ``Full'' and ``None'' represent recognition with each lexicon respectively.}
\label{tab:exp_totaltext}
\end{table}
}
{
\begin{table}
\centering
\begin{tabular}{lccc}
\toprule
\multirow{2}{*}{Method} & \multicolumn{3}{c}{End-to-End} \\
\cmidrule(lr){2-4}
                                            & Strong & Weak & Generic \\
\midrule
SPTS~\cite{peng2022spts}                    & 77.5 & 70.2 & 65.8 \\
\midrule
UNITS$_\textnormal{Shared}$ -- Point        & 89.9 & 84.1 & 79.3 \\
UNITS$_\textnormal{Shared}$ -- Box          & \textbf{90.1} & \textbf{84.5} & 79.3 \\
UNITS$_\textnormal{Shared}$ -- Quad         & 89.9 & \textbf{84.5} & \textbf{79.5} \\
UNITS$_\textnormal{Shared}$ -- Polygon      & 89.4 & 84.0 & 79.0 \\
\bottomrule
\end{tabular}
\caption{The end-to-end recognition performance evaluated by the point-based metric~\cite{peng2022spts} on ICDAR 2015.}
\label{tab:exp_spts_ic15}
\end{table}
}

\subsection{Dataset}

We use the following datasets for the experiments: 
\textbf{Curved SynthText}~\cite{liu2021abcnet} is a synthetic dataset consisting of 94,723 images including mostly straight texts and 54,327 images including curved texts.
It is annotated in bezier annotation formats.
\textbf{ICDAR 2013}~\cite{karatzas2013icdar} is a scene text dataset containing most of the well-captured and horizontal texts, annotated in bounding box formats.
It contains 229 train and 233 test images.
\textbf{ICDAR 2015}~\cite{karatzas2015icdar} is a scene text dataset containing inclined texts, including 1,000 train and 500 test images.
It is annotated in quadrilateral formats.
\textbf{MLT 2019}~\cite{amin2019mlt} is a dataset containing 10,000 train and 10,000 test images.
It contains texts from a total of 7 languages and is annotated in quadrilateral formats.
\textbf{Total-Text}~\cite{ch2017total} is a dataset including arbitrary-shaped texts.
All are annotated in quadrilateral or polygonal formats and it contains 1,255 train and 300 test images.
\textbf{TextOCR}~\cite{singh2021textocr} is a relatively recently published dataset to benchmark text recognition. 
All images are annotated in quadrilateral or polygonal formats, and it is composed of 21,749 train, 3,153 validation, and 3,232 test images.
\textbf{HierText}~\cite{long2022towards} is a most recently published dataset featuring hierarchical annotations of text in natural scenes and documents.
It is annotated in polygonal or quadrilateral annotation formats.
It contains 8,281 train, 1,724 validation, and 1,634 test sets.

\subsection{Implementation Details}

We use Swin-B~\cite{liu2021swin} as the visual encoder of~\model.
The layer numbers and window size are set to \{2, 2, 18, 2\} and 7, respectively.
We use the initial model pre-trained by ImageNet 22k dataset.
For the decoder of~\model, we use 8 transformer decoder layers. 
The maximum length in the decoder is set as 1024, and the decoder weights are randomly initialized.

The model is first pre-trained on a combination of datasets including Curved SynthText, ICDAR 2013, ICDAR 2015, MLT 2019, Total-Text, TextOCR, and HierText.
To ensure the model can learn all detection formats well, 40\% of the annotations are randomly converted into bounding box or central point formats. 
We use a batch size of 128, and the model is pre-trained for 300k steps, with initial 10k warm-up steps.
We use AdamW optimizer with a cosine learning rate scheduler, and set the learning rate as $3e^{-4}$ and the weight decay is set to $1e^{-4}$.
For data augmentation, we apply random rotation, random resize between 30\% and 200\%, random crop, and color jitter. 
The image is resized to 768$\times$768 after being padded in a square shape.
After pre-training, the model is fine-tuned on the same combination of datasets used in pre-training for 120k steps.
We use a batch size of 16 and set the input image resolution as 1920$\times$1920.
In fine-tuning, the AdamW optimizer with cosine learning rate scheduler is used with the learning rate $1.2e^{-4}$.
Additionally, since minor annotation rules are different for each dataset, we fine-tune one more model for each benchmark dataset.
Each model is fine-tuned for 20k steps with a fixed learning rate $3e^{-5}$, and we report the results of unified and dataset-specific models.

\subsection{Results on Text Spotting Benchmarks}

We evaluate the performance of~\model and existing methods on ICDAR 2015 and Total-Text datasets.
For ICDAR 2015, we directly predict quadrilaterals and for Total-Text, we use 16-point polygons.
Both detection and end-to-end results are reported in ~\cref{tab:exp_ic15} and~\cref{tab:exp_totaltext}.
For ICDAR 2015, we use ``strong'', ``weak'', and ``generic'' dictionaries, and also evaluate ``None'' that doesn't use any dictionary.
For Total-Text, we show the results that are ``Full'' with a lexicon and ``None'' without any lexicon.
~\model$_\textnormal{Shared}$ is the unified model, and the other is the model which is fine-tuned on each dataset.

\begin{figure*}
  \centering
  \includegraphics[height=0.35\linewidth]{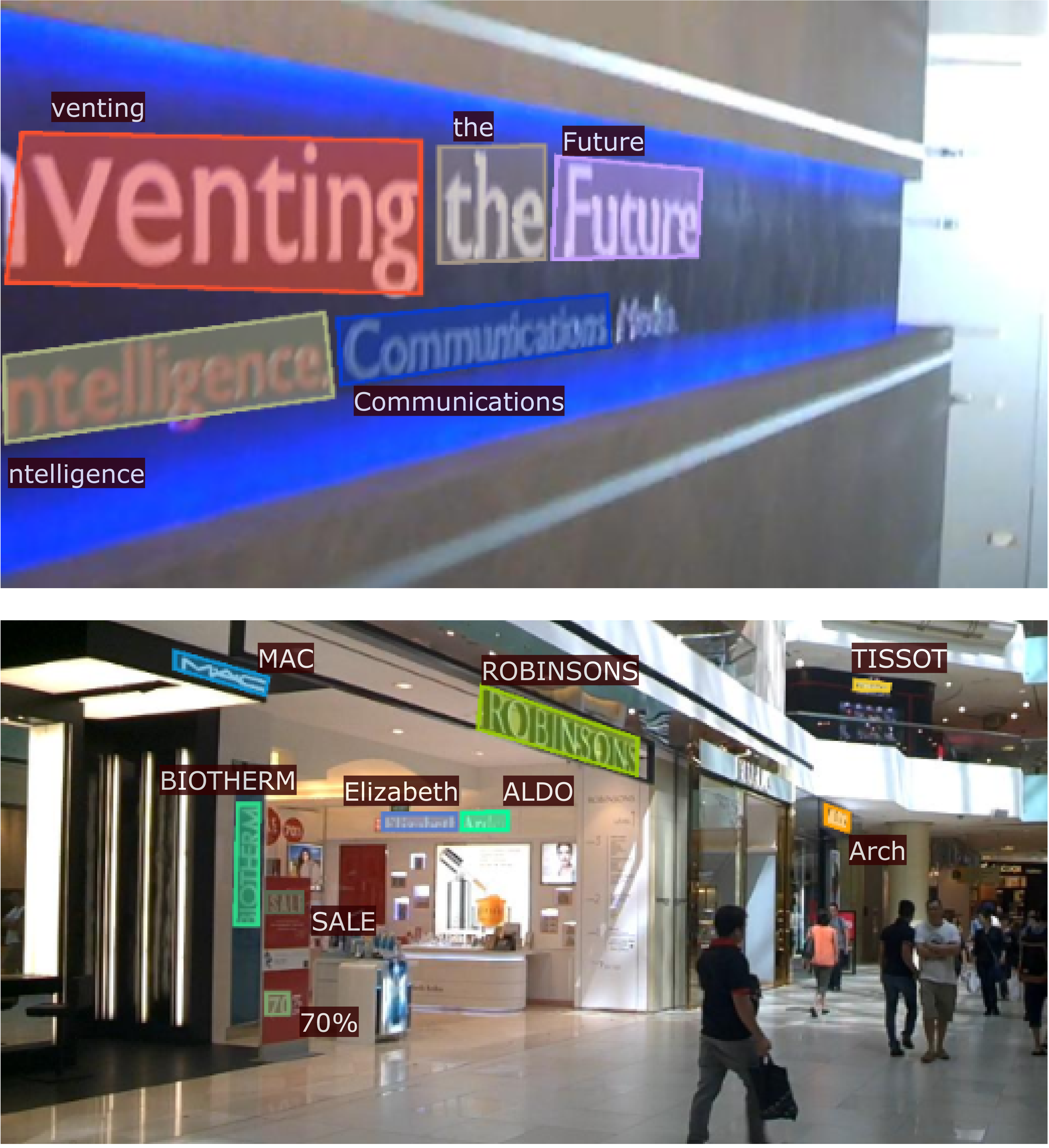}
  \includegraphics[height=0.35\linewidth]{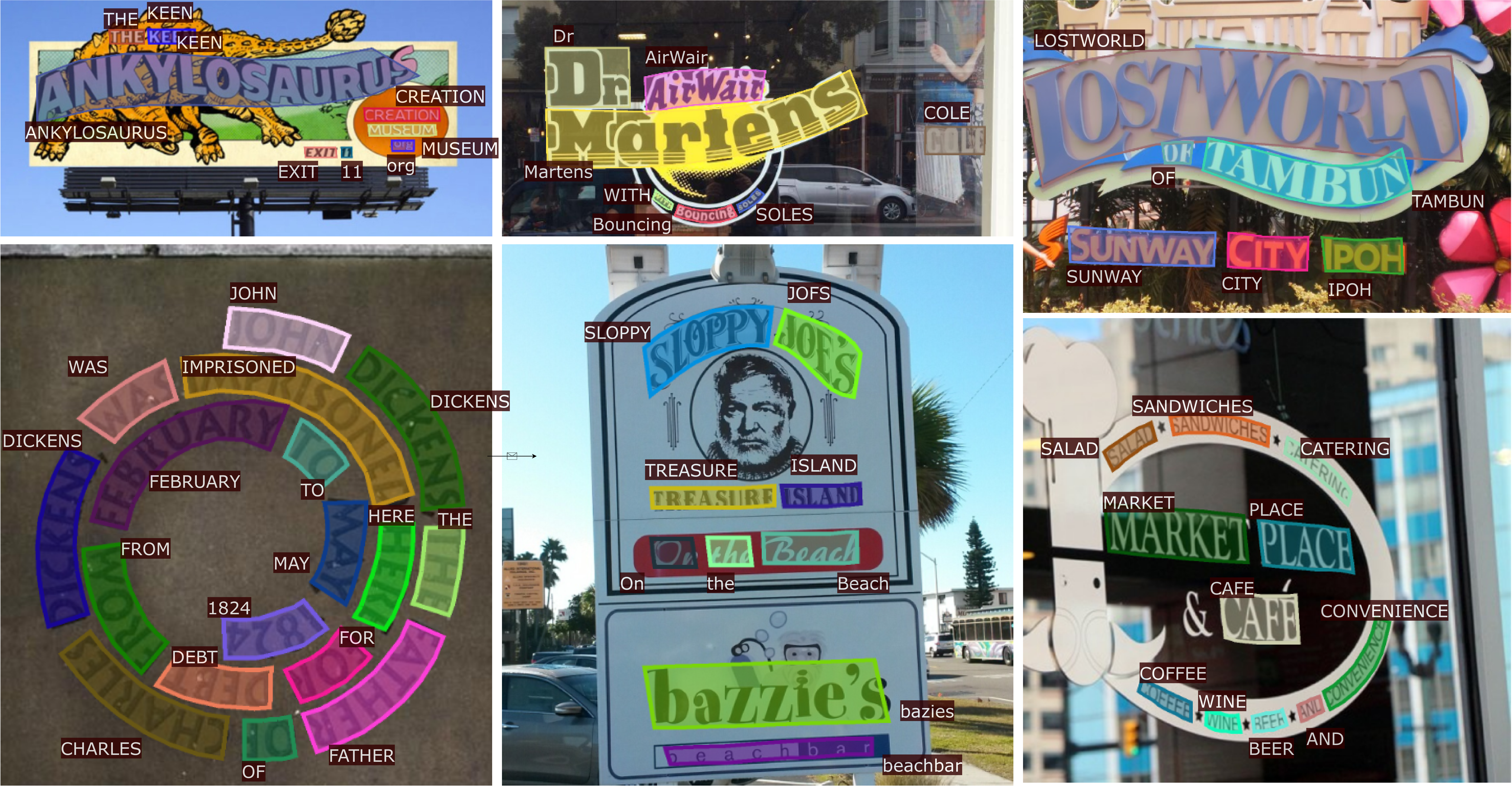}
  \caption{Qualitative results of our method on ICDAR 2015 and Total-Text.}
\label{fig:res_ic15_totaltext}
\end{figure*}

\begin{figure*}
  \centering
  \includegraphics[width=0.24\linewidth]{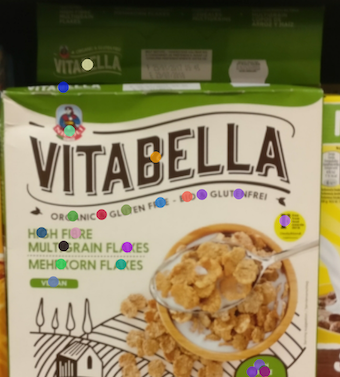}
  \includegraphics[width=0.24\linewidth]{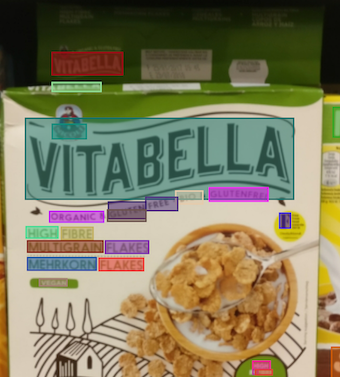}
  \includegraphics[width=0.24\linewidth]{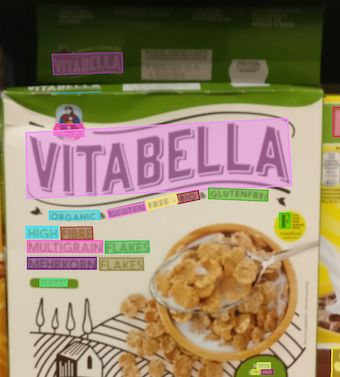}
  \includegraphics[width=0.24\linewidth]{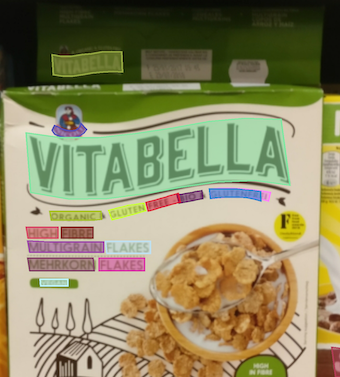}
  \caption{Prediction examples of our unified model. The proposed model can extract text in several detection formats with a single model.}
\label{fig:res_all_formats}
\end{figure*}

On both benchmark datasets, the proposed method shows state-of-the-art performance for all vocabularies in the end-to-end evaluation protocol.
We confirm that our method is a competitive method compared to the existing methods for both quadrilateral and polygonal output formats.
Some qualitative results are shown in~\cref{fig:res_ic15_totaltext,fig:res_all_formats}, where the unified model UNITS$_\textnormal{Shared}$ extracted texts in the desired detection formats.

In addition, we compare the proposed method with the previous study SPTS~\cite{peng2022spts}, which is also a sequence generation method that only handles the single central point format.
SPTS proposed a point-based evaluation metric for evaluating single central point output, and we use the same metric for performance evaluation.
The evaluation results are shown in~\cref{tab:exp_spts_ic15}.
Besides single point output,~\model also can output in the box, quadrilateral, and polygonal formats.
To show the end-to-end spotting performance of all output formats, we calculate the central point for all output formats and measure the performance with the point-based metric.
It shows better performance than the existing SPTS for all detection formats, including central points.
The experimental results for other benchmark datasets are shown in the supplementary material.

\subsection{Ablation studies}

We conduct ablation experiments on ICDAR 2015, Total-Text, and TextOCR to evaluate the effectiveness of the proposed method.
These experiments do not use a lexicon and follow the end-to-end evaluation protocol.

{
\begin{table}
\centering
\begin{tabular}{lcccc}
\toprule
\multirow{2}{*}{Method} & \multicolumn{2}{c}{ICDAR 2015} & \multicolumn{2}{c}{Total-Text} \\
\cmidrule(lr){2-3} \cmidrule(lr){4-5}
                    & DET & E2E & DET & E2E \\
\midrule
Hybrid transformer  & 89.6 & 65.6 & 85.9 & 67.4 \\
Swin transformer    & \textbf{92.0} & \textbf{78.5} & \textbf{88.4} & \textbf{77.3} \\
\bottomrule
\end{tabular}
\caption{Ablation study of the image encoder on ICDAR 2015 and Total-Text. A window attention-based swin transformer capable of high-resolution training and testing shows better performance than the hybrid transformer.}
\label{tab:exp_backbone}
\end{table}
}
{
\begin{table}
\centering
\begin{tabular}{cccc}
\toprule
\multirow{2}{*}{Starting-Point Prompt} & \multicolumn{3}{c}{End-to-End} \\
\cmidrule(lr){2-4}
                & Precision & Recall & F-measure \\
\midrule
-               & 78.4 & 30.6 & 44.0 \\
\checkmark      & \textbf{80.2} & \textbf{54.2} & \textbf{64.7} \\
\bottomrule
\end{tabular}
\caption{Ablation study of the starting-point prompting on TextOCR. The starting-point prompting enables~\model to extract a large number of text instances even with a limited decoder length.}
\label{tab:exp_location_prompting}
\end{table}
}

{
\begin{table}
\centering
\begin{tabular}{cccc}
\toprule
\multirow{2}{*}{Decoder Length} & \multicolumn{3}{c}{End-to-End} \\
\cmidrule(lr){2-4}
                & Precision & Recall & F-measure \\
\midrule
1024            & 80.2 & 54.2 & 64.7 \\
512             & 78.2 & 53.7 & 63.7 \\
256             & 76.8 & 51.8 & 61.8 \\
\bottomrule
\end{tabular}
\caption{Performance of~\model on TextOCR with different decoder lengths. The starting-point prompting works robustly regardless of decoder length, allowing~\model to detect more text instances than the decoder length allows.} 
\label{tab:exp_location_prompting_robustness}
\end{table}
}
{
\begin{table}
\centering
\begin{tabular}{lcccc}
\toprule
\multirow{2}{*}{Method} & \multicolumn{2}{c}{ICDAR 2015} & \multicolumn{2}{c}{Total-Text} \\
\cmidrule(lr){2-3} \cmidrule(lr){4-5}
                    & DET & E2E & DET & E2E \\
\midrule
Vanilla decoder     & 73.7 & 39.1 & 76.1 & 45.1 \\
Multi-way decoder   & \textbf{75.1} & \textbf{40.7} & \textbf{78.9} & \textbf{47.7} \\
\bottomrule
\end{tabular}
\caption{Ablation study of the multi-way decoder on ICDAR 2015 and Total-Text. Both detection and end-to-end performance increase when a multi-way decoder is applied.}
\label{tab:exp_multiway}
\end{table}
}

\noindent\textbf{Image Encoder}.
We compare the swin transformer with the hybrid transformer, which is a combination of ResNet-50~\cite{he2016deep} and vanilla transformer~\cite{vaswani2017attention}, as image encoders.
For the swin transformer, we resize the longer side of the input image to 1920 during both training and inference.
However, the computational complexity of self-attention is quadratic to the image size, which leads to memory issues when training a hybrid transformer at high resolution.
Therefore, we set the longer side of the input image as 1280 when using the hybrid transformer.
The experimental results are shown in the~\cref{tab:exp_backbone}.
In particular, the hybrid transformer shows lower performance.
We confirm that a high-resolution input image is essential for small text detection
and recognition, and for this, the swin transformer, which has linear computational complexity to the image size, is a better choice than a vanilla transformer.

\noindent\textbf{The Starting-Point Prompting}.
The proposed method can detect and recognize more text instances than the maximum number of instances allowed by the decoder length through the starting-point prompting.
To demonstrate this effect, we perform an ablation study by comparing the performance of~\model with and without the starting-point prompt.
We conduct the experiment on the TextOCR dataset, which contains a relatively large number of texts, and evaluate the model using the point-based metric proposed in~\cite{peng2022spts}.
In the experiment, we predict a single central point, so the method without using the prompt is similar to the existing method~\cite{peng2022spts}. The results are shown in~\cref{tab:exp_location_prompting}, where we observe that starting-point prompting significantly improves the recall of the model, as it can detect more instances beyond the maximum sequence lengths. To test the robustness of starting-point prompting, we also evaluate~\model under three different decoder lengths and compare the results. We present the performance in~\cref{tab:exp_location_prompting_robustness} and confirm that the starting-point prompting method works robustly regardless of the decoder length.

\begin{figure}
  \centering
  \begin{subfigure}{.47\linewidth}
    \includegraphics[width=1.0\linewidth]{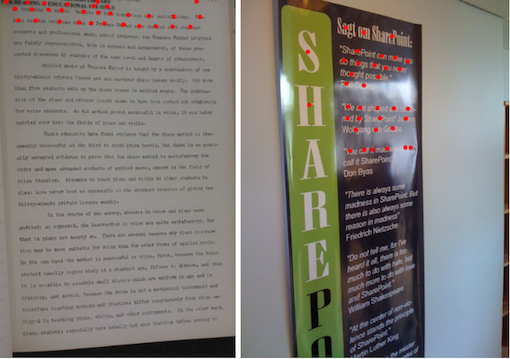}
  \end{subfigure}
  \begin{subfigure}{.47\linewidth}
    \includegraphics[width=1.0\linewidth]{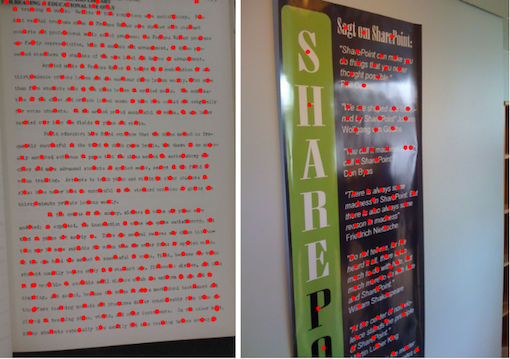}
  \end{subfigure}
  \begin{subfigure}{.47\linewidth}
    \includegraphics[width=1.0\linewidth]{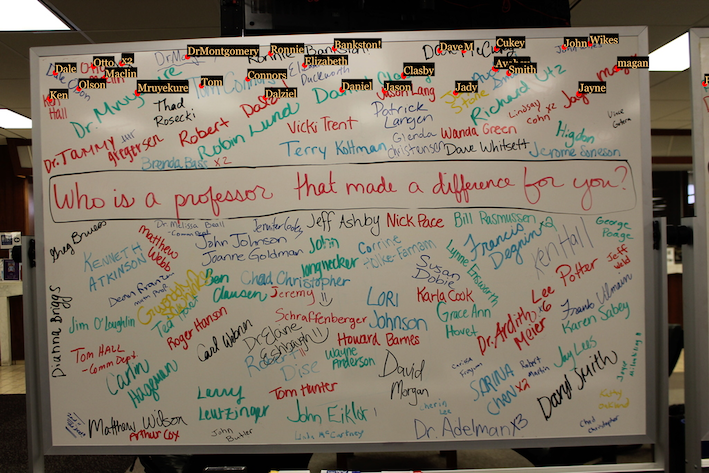}
    \caption{The existing method~\cite{peng2022spts}.}
  \end{subfigure}
  \begin{subfigure}{.47\linewidth}
    \includegraphics[width=1.0\linewidth]{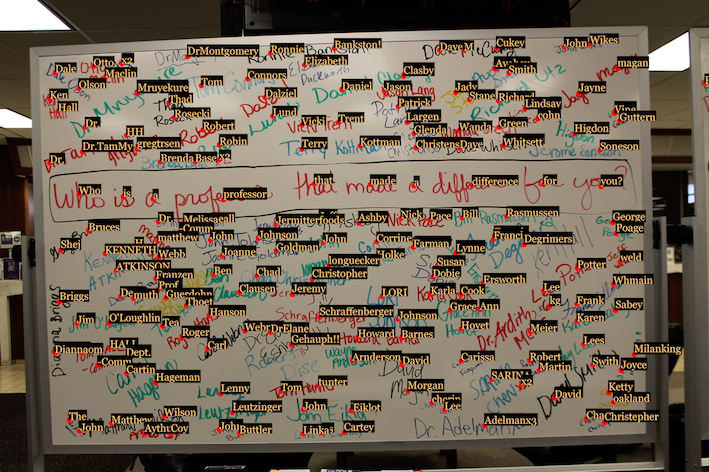}
    \caption{Proposed method.}
  \end{subfigure}
  \caption{Visualization results of the existing sequence generation-based method~\cite{peng2022spts} and our method. Our method overcomes the limitations of existing methods by using the starting-point prompt.}
\label{fig:prompt_ablation}
\end{figure}

As shown in~\cref{fig:prompt_ablation}, starting-point prompting continuously detects and recognizes objects even with a limited decoder length, by taking the starting point as the location of the last detected text instance. In contrast, existing sequence generation-based methods~\cite{peng2022spts} have a limitation in extracting a large number of texts. Our method alleviates this limitation and allows the extraction of texts beyond the number of text boxes that have been trained on.

\noindent\textbf{Multi-way Transformer Decoder}.
We adopt a multi-way transformer decoder structure in which detection and recognition FFN experts are separated to improve the detection and recognition performance of~\model.
To verify the effectiveness of the multi-way transformer decoder structure, we compare it with the vanilla transformer decoder.
In this ablation study, we use a hybrid transformer as a baseline encoder and train both models from scratch with a longer side of the input image of 768, and test them with an image size of 1280.
The results in~\cref{tab:exp_multiway} demonstrate that both the detection and end-to-end performance are improved when the multi-way decoder is applied.
We hypothesize that the multi-way decoder converges faster by determining in advance whether it is the turn to generate a detection token or a recognition token and assigning an expert FFN accordingly.

\section{Conclusions}
In this paper, we have presented a novel end-to-end scene text spotting method that unifies various detection formats to extract arbitrary-shaped text areas.
To handle multiple detection formats simultaneously, we employ a multi-way transformer decoder, and the starting-point prompting strategy enables our model to extract a larger number of texts than seen in training data.
Experimental results show that our method achieves competitive results on benchmark datasets, outperforming text-spotting-specific models even with a simple sequence generation model.
The success of our method suggests its potential applicability in other vision tasks, such as object detection or segmentation.

\clearpage

{\small
\bibliographystyle{ieee_fullname}
\bibliography{egbib}
}

\newpage
\clearpage
\appendix

\renewcommand{\thefigure}{A\arabic{figure}}
\renewcommand{\thetable}{A\arabic{table}}
\setcounter{figure}{0}
\setcounter{table}{0}
\section{Implementation Details}
\label{sec:implementation}
{
\begin{table*}
\centering
\begin{tabular}{lcccccccc}
\toprule
\multirow{2}{*}{Method}                 & \multicolumn{4}{c}{Word Spotting} & \multicolumn{4}{c}{End-to-End}    \\
\cmidrule(lr){2-5} \cmidrule(lr){6-9}
                                        & Strong & Weak & Generic & None & Strong & Weak & Generic & None \\
\midrule
CRAFTS~\cite{baek2020character}         & - & - & - & - & 83.1 & 82.1 & 74.9 & - \\
MaskTextSpotter v3~\cite{liao2020mask}  & 83.1 & 79.1 & 75.1 & - & 83.3 & 78.1 & 74.2 & - \\
ABCNet v2~\cite{liu2020abcnet}          & - & - & - & - & 82.7 & 78.5 & 73.0 & - \\
MANGO~\cite{qiao2021mango}              & 85.2 & 81.1 & 74.6 & - & 85.4 & 80.1 & 73.9 & - \\
DEER~\cite{kim2022deer}                 & - & - & - & - & 82.7 & 79.1 & 75.6 & 71.7 \\
SwinTextSpotter~\cite{huang2022swintextspotter} & - & - & - & - & 83.9 & 77.3 & 70.5 & - \\
TESTR~\cite{zhang2022text}              & - & - & - & - & 85.2 & 79.4 & 73.6 & 65.3 \\
TTS~\cite{kittenplon2022towards}        & 85.0 & 81.5 & 77.3 & - & 85.2 & 81.7 & 77.4 & - \\
GLASS~\cite{ronen2022glass}             & 86.8 & 82.5 & 78.8 & - & 85.3 & 79.8 & 74.0 & - \\
\midrule
UNITS$_\textnormal{Shared}$             & \underline{88.1}  & \underline{84.9}  & \underline{80.7}  & \underline{78.7}  & \underline{88.4}  & \underline{83.9}  & \underline{79.7}  & \underline{78.5}\\
UNITS                                   & \textbf{88.8}  & \textbf{85.2}  & \textbf{81.5}  & \textbf{78.8}  & \textbf{89.0}  & \textbf{84.1}  & \textbf{80.3}  & \textbf{78.7}\\
\bottomrule
\end{tabular}
\caption{Experiment results on ICDAR 2015. ``Strong'', ``Weak'', ``Generic'' and ``None'' represent recognition with each lexicon respectively.}
\label{tab:exp_suppl_ic15}
\end{table*}
}

\begin{figure*}
  \centering
    \includegraphics[width=1.0\linewidth]{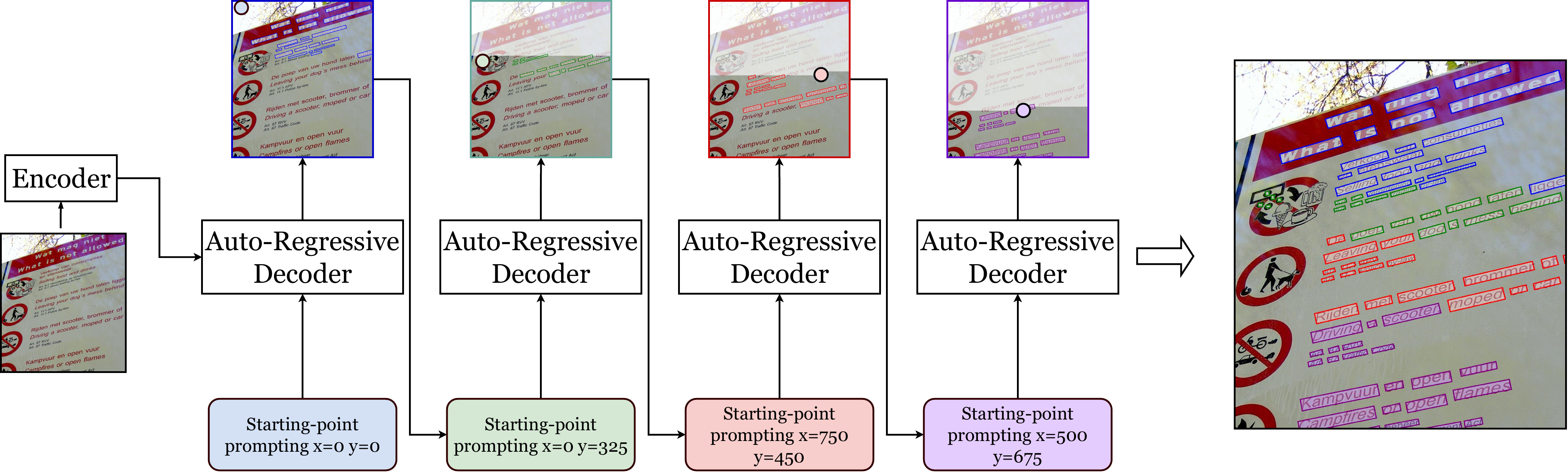}
    \hfill
  \caption{Illustration of text extraction using the starting-point prompt.}
\label{fig:starting_prompt_test_detail}
\end{figure*}

\begin{figure}
  \centering
    \includegraphics[width=1.0\linewidth]{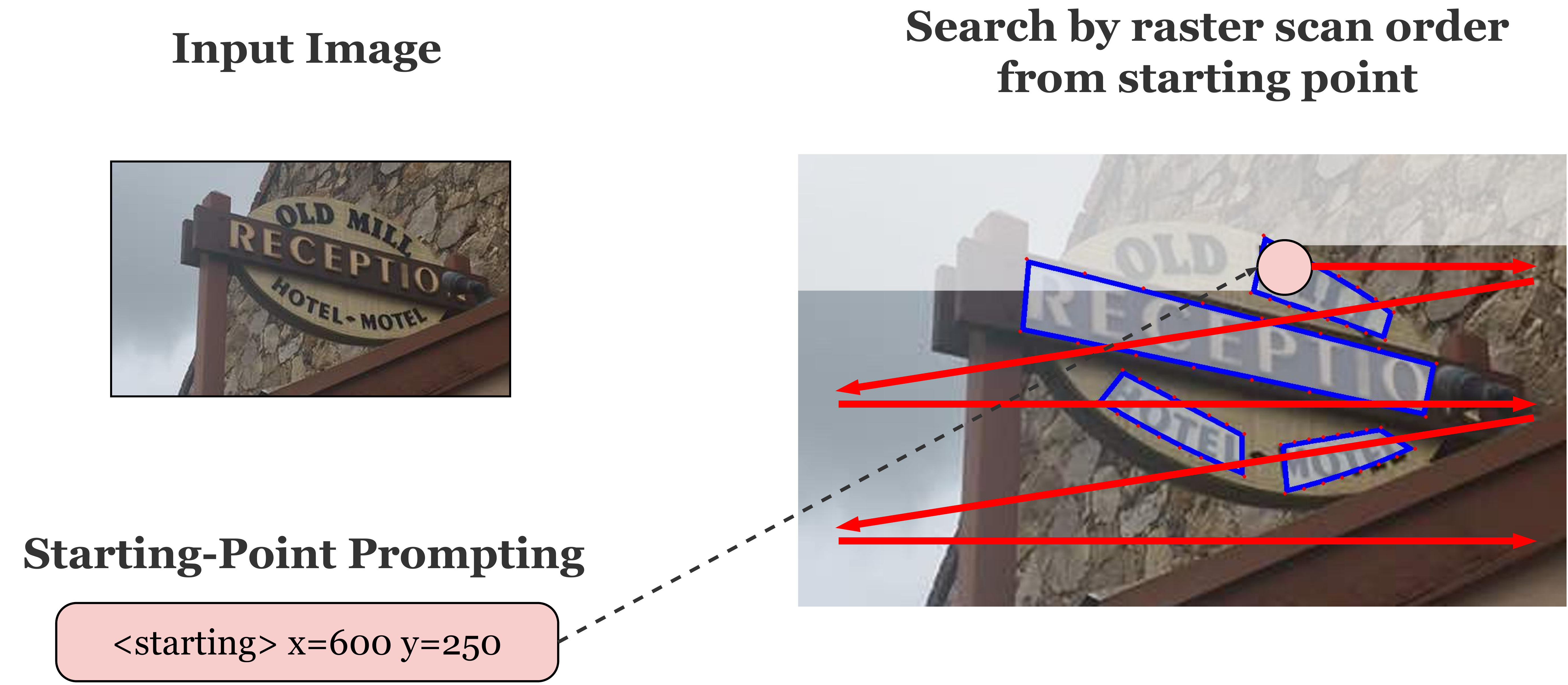}
    \hfill
  \caption{Illustration of the order of text extraction.}
\label{fig:prompt_search}
\end{figure}

\subsection{The Starting-Point Prompting}
Our proposed method applies starting-point prompting to enable the model to extract texts from an arbitrary starting point, allowing it to generate a longer sequence than the maximum decoding length.
To apply starting-point prompting, we use raster scan order as the order of text instance extraction.
The text extraction process is shown in~\cref{fig:starting_prompt_test_detail}.
Text instances with central points within the search region are extracted in raster scan order.
Furthermore, during testing, if the generated output sequence does not end with an $<$eos$>$ token, our method sets the starting point as the last detected text position in the previous step, then continues and enables the re-generation of a text sequence corresponding to the remaining objects, as shown in~\cref{fig:prompt_search}.

\subsection{Multi-way Decoder}
\begin{figure}
  \centering
  \begin{subfigure}{.49\linewidth}
    \includegraphics[width=1.0\linewidth]{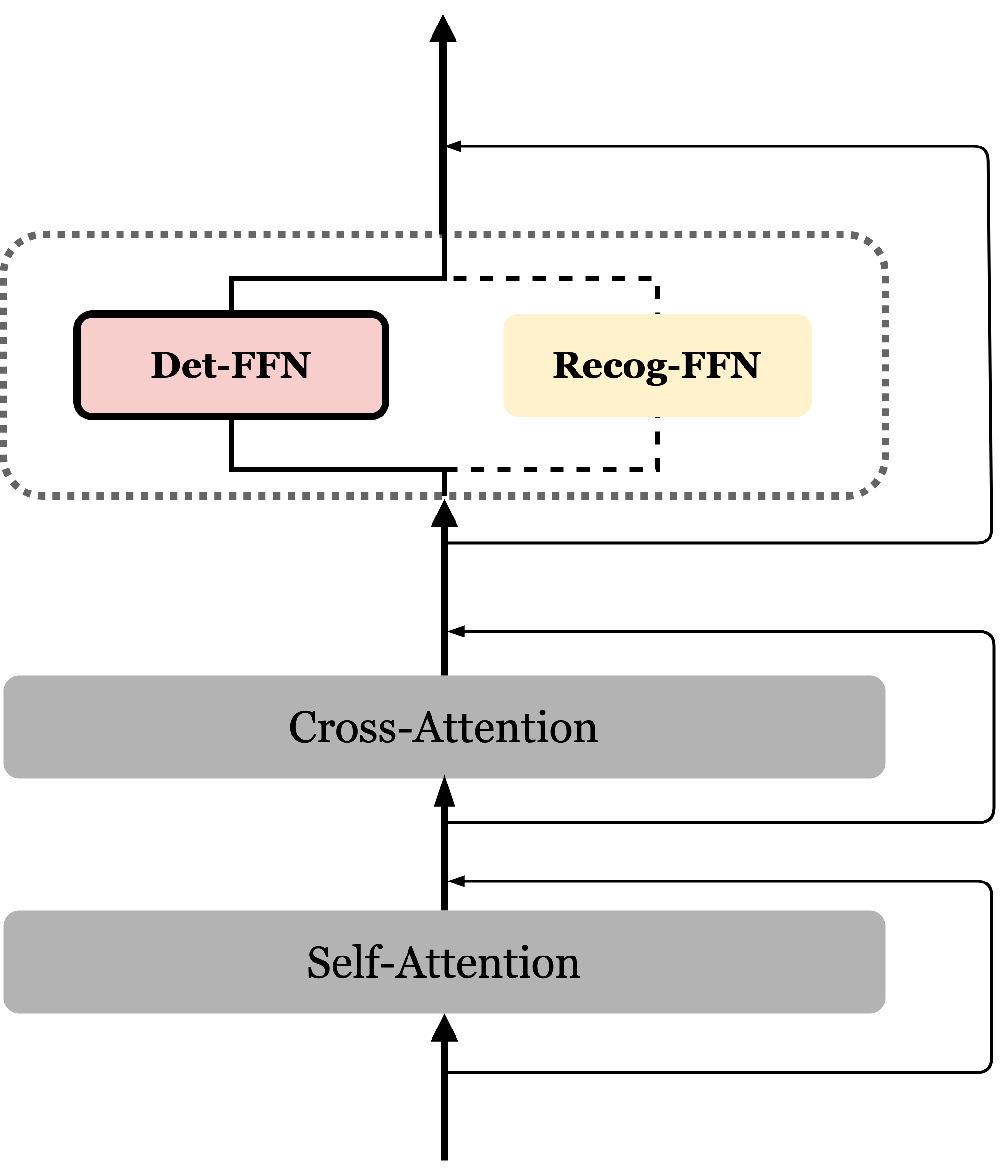}
    \caption{Figure of the decoder when generating a detection token.}
  \end{subfigure}
  \begin{subfigure}{.49\linewidth}
    \includegraphics[width=1.0\linewidth]{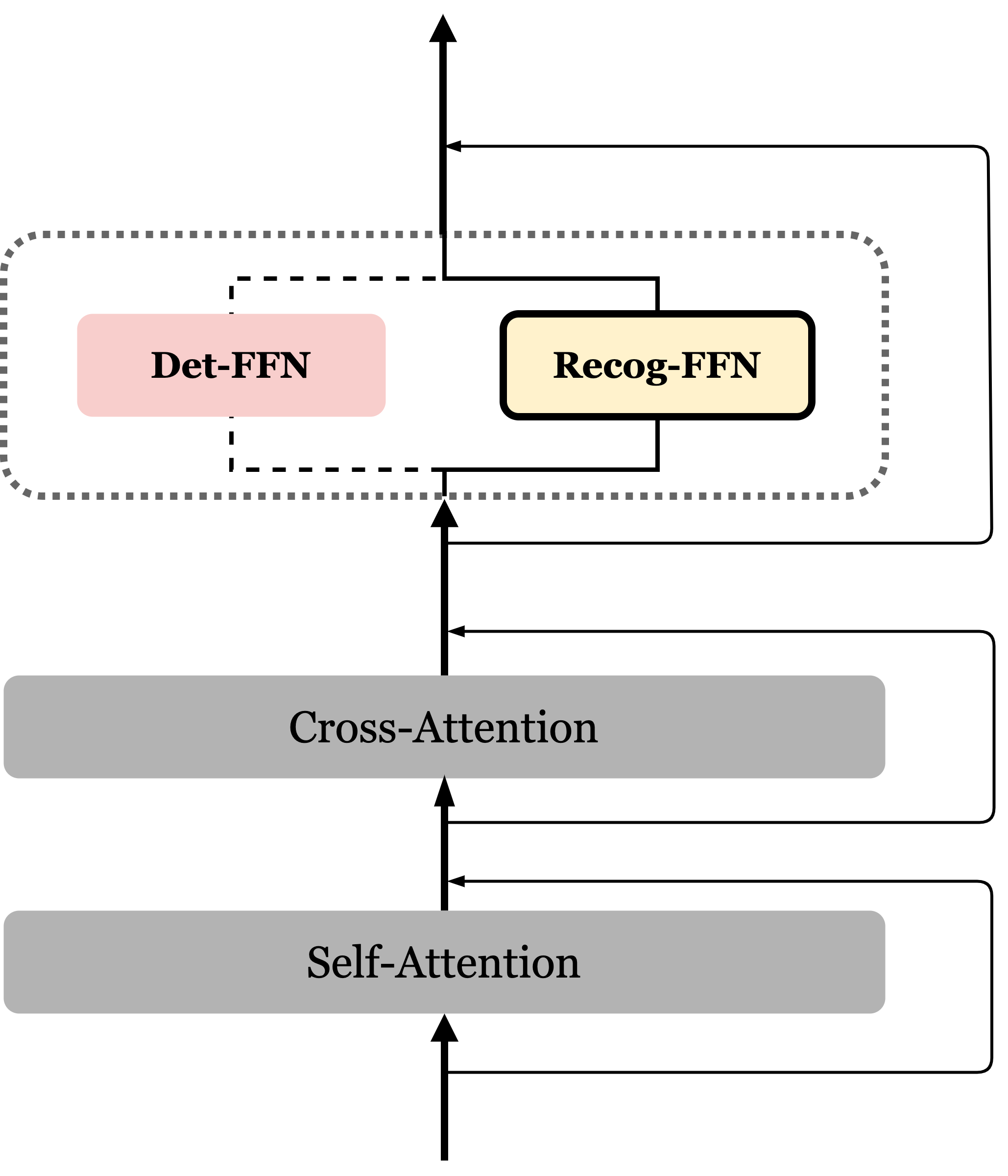}
    \caption{Figure of the decoder when generating a recognition token.}
  \end{subfigure}
  \caption{Illustration of the multi-way transformer decoder.}
\label{fig:multiway}
\end{figure}

Our model employs a multi-way transformer decoder, as shown in~\cref{fig:multiway}.
The detection and recognition feedforward networks (FFNs) are separated, and attention layers are shared between them.
When generating a detection token, the model is required to pass it through the detection FFN, and when it is time to generate a recognition token, the model must pass it through the recognition FFN.
This separation enables the model to better learn multiple detection formats simultaneously, and the shared attention modules can learn from both tasks, improving the overall performance.

\section{More Experiments}
\label{sec:experiments}
{
\begin{table}
\centering
\begin{tabular}{lcccc}
\toprule
\multirow{2}{*}{Method}                 & \multicolumn{2}{c}{Word Spotting} & \multicolumn{2}{c}{End-to-End}    \\
\cmidrule(lr){2-3} \cmidrule(lr){4-5}
                                        & None & Full & None & Full \\
\midrule
CRAFTS~\cite{baek2020character}         & - & - & \textbf{78.7} & - \\
MaskTextSpotter v3~\cite{liao2020mask}  & 75.1 & 81.8  & 71.2 & 78.4 \\
ABCNet v2~\cite{liu2020abcnet}          & - & - & 70.4 & 78.1 \\
MANGO~\cite{qiao2021mango}              & 72.9 & 83.6 & 68.9 & 78.9 \\
DEER~\cite{kim2022deer}                 & - & - & 74.8 & 83.3 \\
SwinTextSpotter~\cite{huang2022swintextspotter} & - & - & 74.3 & 84.1 \\
TESTR~\cite{zhang2022text}              & - & - & 73.3 & 83.9 \\
TTS~\cite{kittenplon2022towards}        & 78.2 & 86.3 & 75.6 & 84.4 \\
GLASS~\cite{ronen2022glass}             & 79.9 & 86.2 & 76.6 & 83.0 \\
\midrule
UNITS$_\textnormal{Shared}$                                   & \underline{81.2}  & \underline{87.0} & 77.3  & \underline{85.0}  \\
UNITS                                & \textbf{82.2}  & \textbf{88.0} & \textbf{78.7}  & \textbf{86.0}  \\
\bottomrule
\end{tabular}
\caption{Experiment results on Total-Text. ``Full'' and ``None'' represent recognition with each lexicon respectively.}
\label{tab:exp_suppl_totaltext}
\end{table}
}
{
\begin{table}
\centering
\begin{tabular}{lccc}
\toprule
\multirow{2}{*}{Method}                 & Detection & \multicolumn{2}{c}{End-to-End}    \\
\cmidrule(lr){2-2} \cmidrule(lr){3-4}
                                        & F-measure & None & Full \\
\midrule
ABCNet v2~\cite{liu2020abcnet}          & 84.7 & 51.8 & 77.0 \\
MANGO~\cite{qiao2021mango}              & - & \underline{58.9} & 78.7 \\
SwinTextSpotter~\cite{huang2022swintextspotter} & \underline{88.0} & 51.8 & 77.0 \\
TESTR~\cite{zhang2022text}              & 87.1 & 56.0 & \underline{81.5} \\
\midrule
UNITS                                   & \textbf{88.6}  & \textbf{66.4} &  \textbf{82.3} \\
\bottomrule
\end{tabular}
\caption{Experiment results on CTW1500. ``Full'' and ``None'' represent recognition with each lexicon respectively.}
\label{tab:exp_suppl_ctw}
\end{table}
}
{
\begin{table}
\centering
\begin{tabular}{lccc}
\toprule
\multirow{2}{*}{Method}                 & \multicolumn{3}{c}{End-to-End}    \\
\cmidrule(lr){2-4}
                                        & Strong & Weak & Generic \\
\midrule
CRAFTS~\cite{baek2020character}         & 94.2 & 93.8 & 92.2 \\
MaskTextSpotter v2~\cite{lyu2018mask}   & 93.3 & 91.3 & 88.2 \\
MANGO~\cite{qiao2021mango}              & 93.4 & 92.3 & 88.7 \\
\midrule
UNITS$_\textnormal{Shared}$  -- Box     & \textbf{95.1}  & \textbf{94.6}  & \underline{92.9} \\
UNITS$_\textnormal{Shared}$  -- Quad    & \textbf{95.1} & \textbf{94.6} & \textbf{93.0} \\
\bottomrule
\end{tabular}
\caption{Experiment results on ICDAR 2013. ``Strong'', ``Weak'', and ``Generic'' represent recognition with each lexicon respectively.}
\label{tab:exp_suppl_ic13}
\end{table}
}
{
\begin{table}
\centering
\begin{tabular}{lcccc}
\toprule
\multirow{2}{*}{Method}                 & \multicolumn{2}{c}{45\textdegree{}} & \multicolumn{2}{c}{60\textdegree{}}    \\
\cmidrule(lr){2-3} \cmidrule(lr){4-5}
                                        & DET & E2E & DET & E2E \\
\midrule
MaskTextSpotter v3~\cite{liao2020mask}  & 84.2 & 76.1 & 84.7 & 76.6 \\
SwinTextSpotter~\cite{huang2022swintextspotter} & - & 77.6 & - & 77.9  \\
TTS~\cite{kittenplon2022towards}        & \underline{88.8} & \underline{80.4} & \underline{87.6} & \textbf{80.1}  \\
\midrule
UNITS$_\textnormal{Shared}$             & \textbf{91.8} & \textbf{80.6} & \textbf{90.3} & \underline{78.1} \\
\bottomrule
\end{tabular}
\caption{Experiment results on Rotated ICDAR 2013. The end-to-end recognition task is evaluated without any lexicon.}
\label{tab:exp_suppl_roc13}
\end{table}
}
{
\begin{table}
\centering
\begin{tabular}{lcccccccc}
\toprule
\multirow{2}{*}{Method}                 & \multicolumn{3}{c}{Detection}  & \multirow{2}{*}{End-to-End}   \\
\cmidrule(lr){2-4}
                                        & R & P & F &  \\
\midrule
UNITS$_\textnormal{Shared}$             & 63.1 & 83.2 & 71.8 & 61.4 \\
UNITS                                   & 67.1 & 84.8 & 74.9 & 63.6 \\
\bottomrule
\end{tabular}
\caption{Experiment results on TextOCR validation sets. ``R'', ``P'', and ``F'' represent recall, precision and F-measure respectively.}
\label{tab:exp_suppl_textocr}
\end{table}
}

\subsection{Word Spotting Evaluation on Benchmarks}
In the main paper, we reported the end-to-end evaluation scores on the ICDAR 2015 and Total-Text datasets.
In text spotting, there are two evaluation protocols: end-to-end and word-spotting.
In this section, we report both word spotting and end-to-end scores on these two benchmark datasets in~\cref{tab:exp_suppl_ic15,tab:exp_suppl_totaltext}.
End-to-end results are repeated for comparison with word spotting.
We confirm that the proposed method shows state-of-the-art performance for all vocabularies in both evaluation protocols.

\subsection{Evaluation on CTW-1500, ICDAR 2013, and Rotated ICDAR 2013}
\begin{figure}
  \centering
    \includegraphics[width=1.0\linewidth]{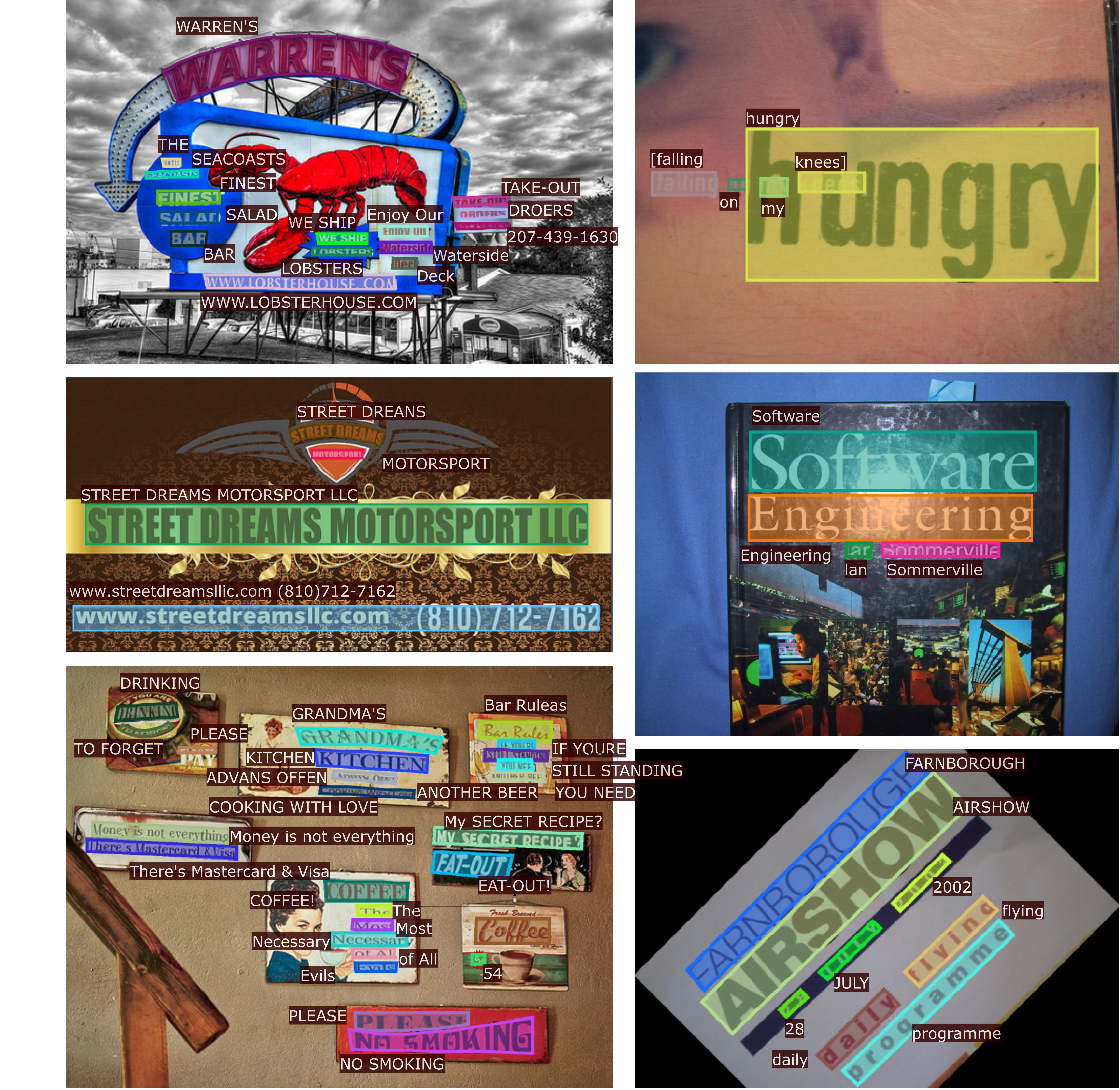}
    \hfill
  \caption{Qualitative results of our method on CTW-1500, ICDAR 2013 and Rotated ICDAR 2013.}
\label{fig:res_suppl}
\end{figure}

To demonstrate the superiority of our method, further evaluation is conducted on other benchmark datasets: CTW-1500~\cite{liu2019curved}, ICDAR 2013~\cite{karatzas2013icdar}, and Rotated ICDAR 2013~\cite{liao2020mask}.
CTW-1500 includes arbitrary-shaped texts and contains 1,000 training and 500 testing images.
Unlike the other datasets, CTW-1500 is annotated in not word-level but text line-level annotations.
Since this setting is different from the unified model learned with word-level annotation, we fine-tune the model with the CTW-1500 dataset and only measure the performance of the fine-tuned model.
In fine-tuning, we set the maximum length of each text transcription to 100.
The model is fine-tuned for 20k steps with fixed learning rate $3e^{-5}$ from unified model UNITS$_\textnormal{Shared}$.
For evaluating CTW-1500, we predict 16-point polygons, and for ICDAR 2013, we use both bounding box and quadrilaterals.
For Rotated ICDAR 2013, we use 4-point quadrilaterals.
~\cref{tab:exp_suppl_ctw,tab:exp_suppl_ic13,tab:exp_suppl_roc13} show that our method achieves competitive results with the existing methods.
Some qualitative results are shown in~\cref{fig:res_suppl}.

\subsection{Evaluation on TextOCR}
\begin{figure*}
  \centering
  \begin{subfigure}{.49\linewidth}
    \includegraphics[width=1.0\linewidth]{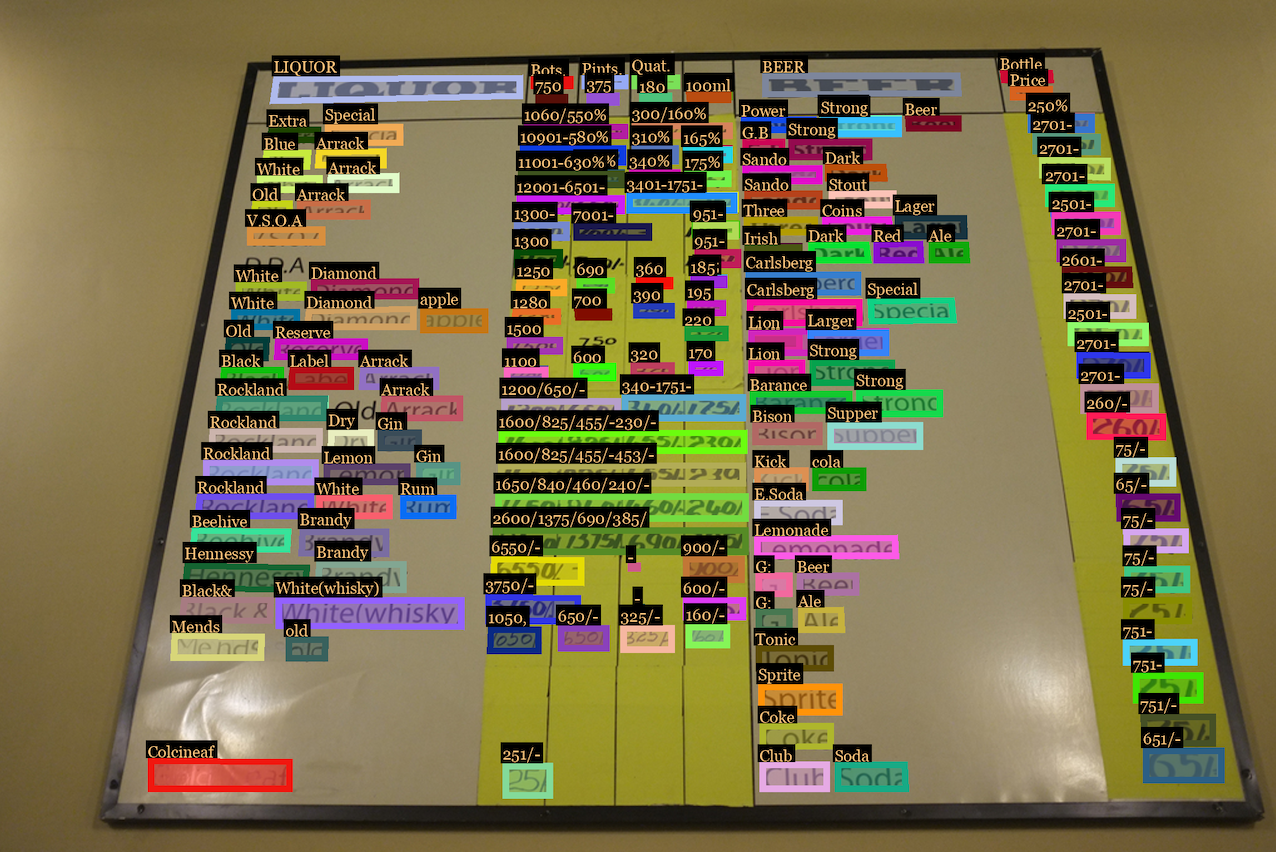}
  \end{subfigure}
  \begin{subfigure}{.49\linewidth}
    \includegraphics[width=1.0\linewidth]{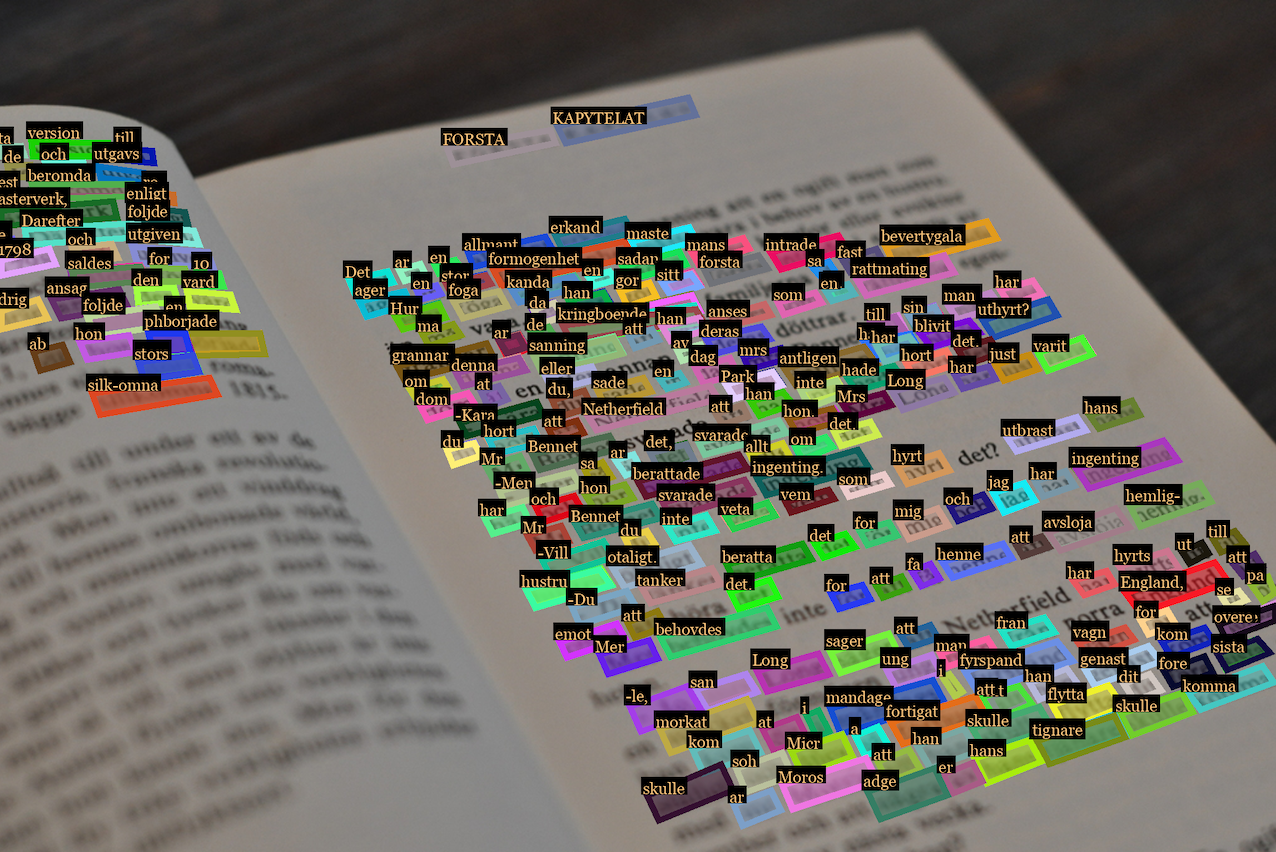}
  \end{subfigure}
    \begin{subfigure}{.49\linewidth}
    \includegraphics[width=1.0\linewidth]{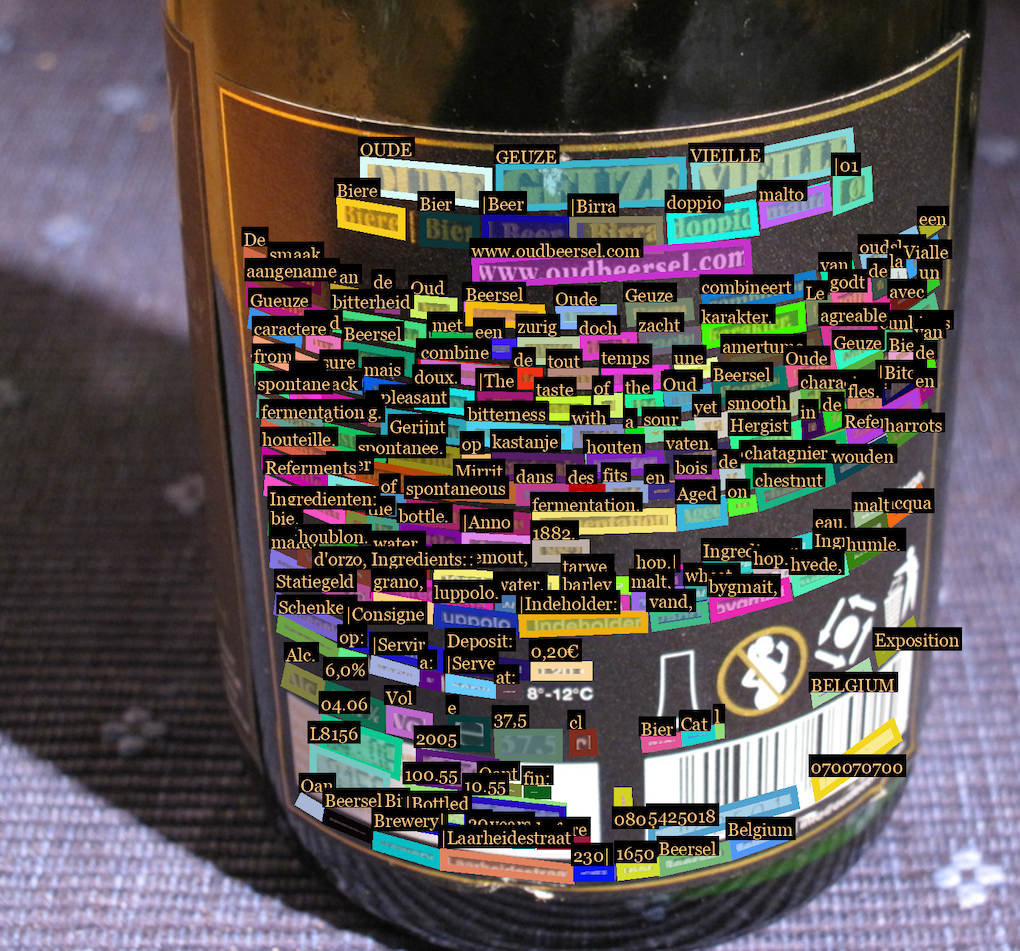}
  \end{subfigure}
  \begin{subfigure}{.49\linewidth}
    \includegraphics[width=1.0\linewidth]{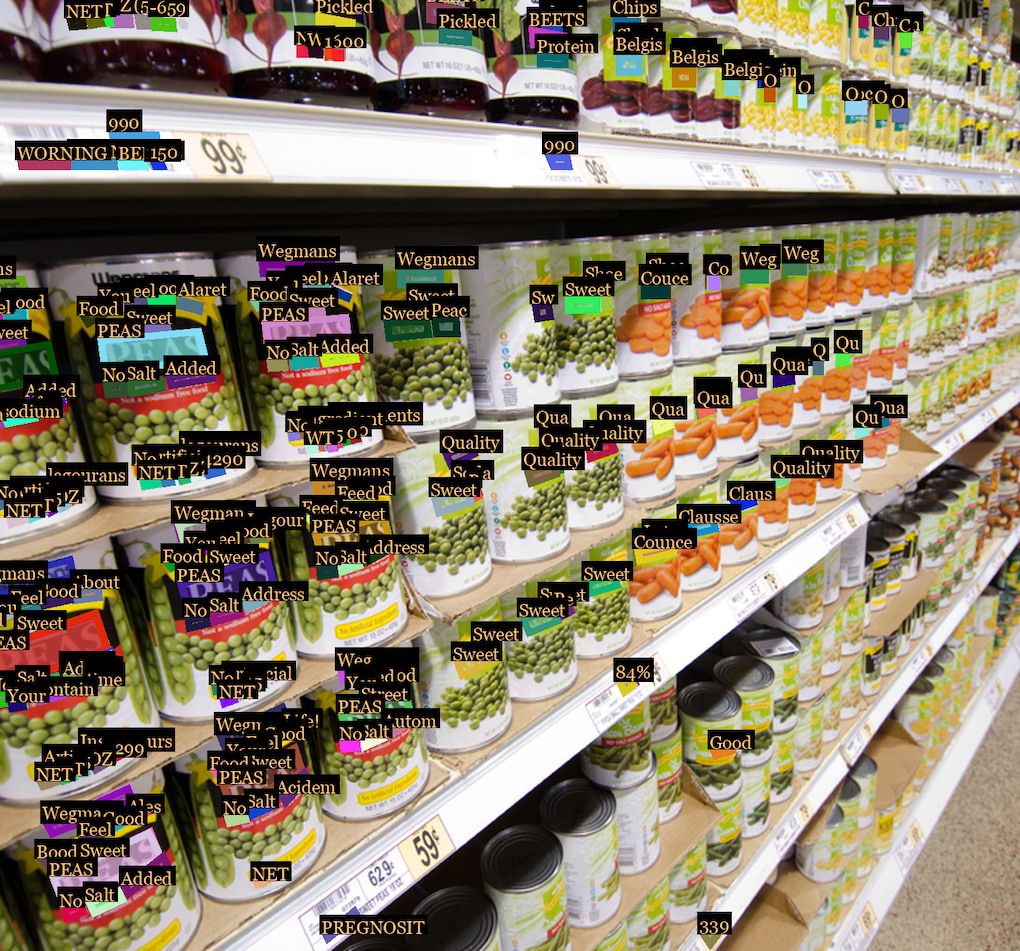}
  \end{subfigure}
  \caption{
  Qualitative results of our method on TextOCR.
  }
\label{fig:res_suppl_textocr1}
\end{figure*}

\begin{figure*}
  \centering
  \includegraphics[width=0.98\linewidth]{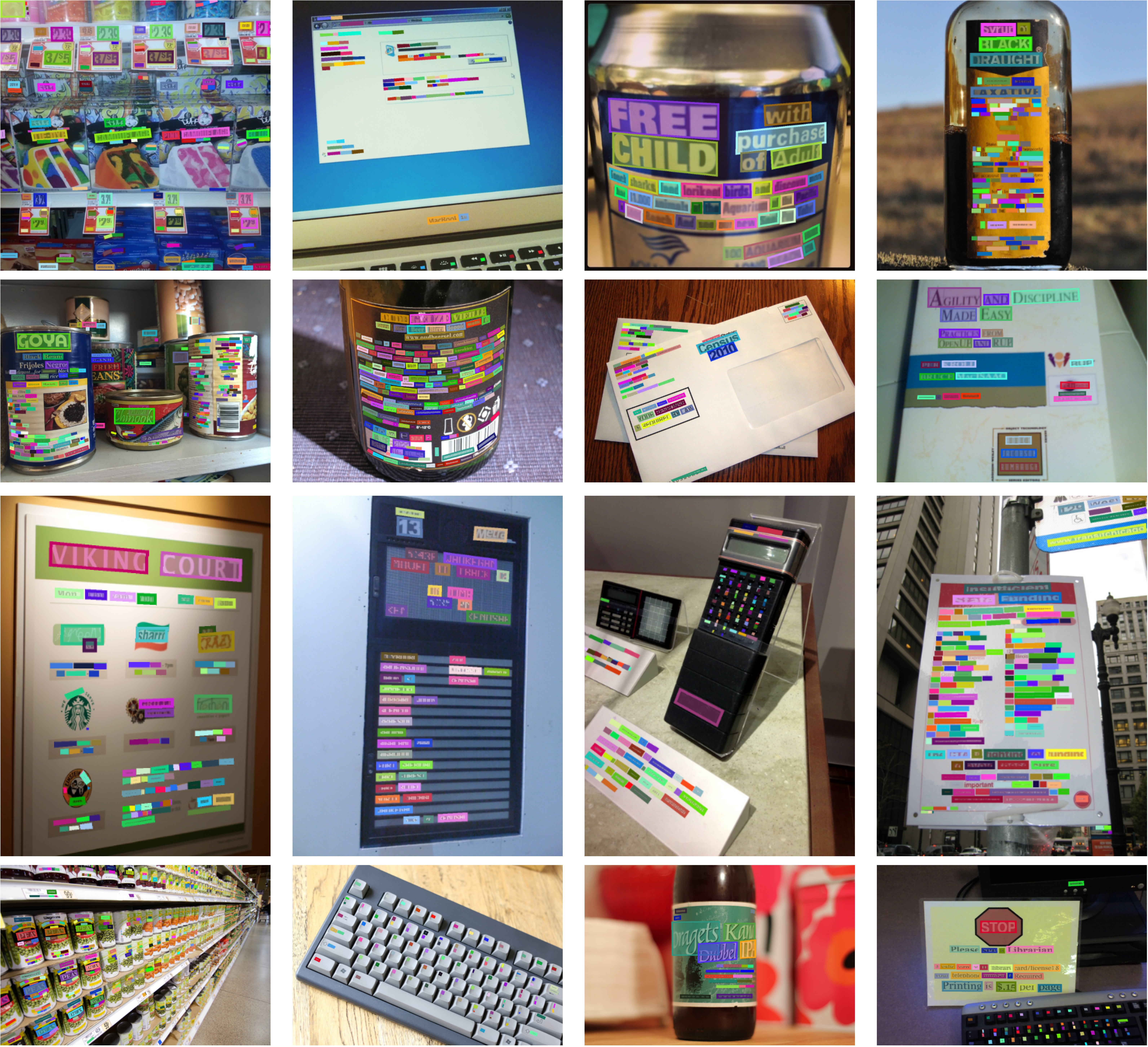}
  \caption{Qualitative detection results of our method on TextOCR.}
\label{fig:res_suppl_textocr2}
\end{figure*}

The proposed method can detect and recognize more text instances than the maximum number of instances allowed by the decoder length.
To demonstrate this effect, we evaluate on TextOCR dataset, which contains a relatively large number of texts.
Since TextOCR does not provide annotations for the test sets, the performance evaluation is performed for validation sets.
We did not use validation sets at all in training or model selection.
Similar to the experiments in the main paper, we fine-tune the model for TextOCR and report the results of both the unified and fine-tuned models.
The model is fine-tuned for 150k steps with a fixed learning rate of $3e^{-5}$ from the unified model UNITS$_\textnormal{Shared}$. 
The detection and end-to-end scores are reported in~\cref{tab:exp_suppl_textocr}.
Some qualitative results are shown in~\cref{fig:res_suppl_textocr1,fig:res_suppl_textocr2}

\end{document}